%% file: main_full.tex
\pdfoutput=1

\documentclass[dvipsnames,format=sigconf,anonymous=false,authorversion,nonacm]{acmart}

\usepackage{color,soul}
\usepackage{lipsum}
\usepackage{hyperref}
\usepackage{enumitem}
\usepackage{caption}
\usepackage{subcaption}
\usepackage{cleveref}

\usepackage{adjustbox}
\usepackage[detect-weight=true, binary-units=true]{siunitx}
\usepackage{multicol,multirow,booktabs,makecell,array}
\usepackage[linesnumbered]{algorithm2e}
\crefname{algocf}{algorithm}{algorithms}
\Crefname{algocf}{Algorithm}{Algorithms}
\usepackage{gensymb}

\usepackage{tikz,pgfplots,pgfplotstable}
\usetikzlibrary{patterns}
\pgfplotsset{compat=1.17}
\usepgfplotslibrary{groupplots,fillbetween,statistics}
\usetikzlibrary{external,decorations.pathreplacing,calligraphy,shapes,arrows,fit}
\tikzset{external/only named=true}
\usepackage[basecolor=blue!60,size=3mm]{vsrs}
\input{drawing}

\input{plot-macros}

\AtBeginDocument{%
 \providecommand\BibTeX{{%
 \normalfont B\kern-0.5em{\scshape i\kern-0.25em b}\kern-0.8em\TeX}}}

\begin{document}

\title{Self-building Neural Networks}

\author{Andrea Ferigo}
\author{Giovanni Iacca}

\affiliation{%
 \institution{University of Trento}
 \streetaddress{Via Sommarive, 9}
 \city{Trento}
 \country{Italy}
 \postcode{38123}
}

\renewcommand{\shortauthors}{Ferigo and Iacca}




\begin{abstract}
During the first part of life, the brain develops while it learns through a process called synaptogenesis. The neurons, growing and interacting with each other, create synapses. However, eventually the brain prunes those synapses. While previous work focused on learning and pruning independently, in this work we propose a biologically plausible model that, thanks to a combination of Hebbian learning and pruning, aims to simulate the synaptogenesis process. In this way, while learning how to solve the task, the agent translates its experience into a particular network structure. Namely, the network structure \emph{builds itself} during the execution of the task. We call this approach \emph{Self-building Neural Network} (SBNN). We compare our proposed SBNN with traditional neural networks (NNs) over three classical control tasks from OpenAI. The results show that our model performs generally better than traditional NNs. Moreover, we observe that the performance decay while increasing the pruning rate is smaller in our model than with NNs. Finally, we perform a validation test, testing the models over tasks unseen during the learning phase. In this case, the results show that SBNNs can adapt to new tasks better than the traditional NNs, especially when over $80\%$ of the weights are pruned.
\end{abstract}

\maketitle


\section{Introduction}
\label{sec:intro}
The natural brain is one of the most complex systems we know of. It can perform complex tasks and adapt to new situations with an efficiency that is currently unreachable by any modern Artificial Intelligence (AI) system. 
These performances derive from a long-lasting evolutionary process that has harmonized a vast amount of different elements that work at different scales.
At the molecular level, for example, some cells have to provide energy to other cells by exchanging Adenosine triphosphate (ATP). Other molecules, instead, are used to regulate inter-cellular communication.
At the cellular level, stem cells have to specialize into glial or neuron cells. Eventually, these cells grow and create connections, composing our brain.
This process, called synaptogenesis~\cite{ZUKERMAN20119}, is then followed by a second process that prunes the less relevant synapses~\cite{bourgeois_2010, Tau2010NormalDO}. It is believed that this pruning phase is one of the main reasons why brains are such efficient systems~\cite{CHUNG2012438, PRESUMEY201753,lichtman2000synapse}.

In this work, we focus on this last aspect: the growth and organization of the neurons in the brain.
Specifically, we are interested in linking the growth and organization to the experience of the agent during the task, as it happens in the natural brain.
To simulate this process, we combine two well-known mechanisms from the literature.

Firstly, we consider a Neural Network (NN), on top of which we apply a synaptic plasticity mechanism, which changes the synapses of the NN based on the experience of the agent during the task. In this case, we adopt Hebbian Learning~\cite{brown1990hebbian}, a task-agnostic plasticity model that takes inspiration from the natural neurons. 

Secondly, we include in our model a pruning mechanism based on the global magnitude algorithm \cite{Bishop1995NeuralNF,han2015pruning}. However, we modify this algorithm in such a way to decide not only \emph{how much} (i.e., how many synapses) to prune, but also \emph{when} prune.

We call the resulting model \emph{Self-building Neural Network} (SBNN), due to its capability to compose its own structure based on the experience perceived by the agent during its life. 

We test our proposed SBNN in three classical control tasks from OpenAI, to show the capability of this model in terms of performance and how the structure of the networks can be different depending on the task. In a separate set of experiments, we also assess the generalization capabilities of the SBNN. 

The rest of this paper is organized as follows. \Cref{sec:rw} summarizes the related works. \Cref{sec:methodology} describes the methods, and in particular the proposed SBNN.
Then, \Cref{sec:results} shows the results, followed by the conclusions in \Cref{sec:conclusions}.


\section{Related work}
\label{sec:rw}
As described in the previous section, our SBNN is based on two mechanisms: plasticity and pruning.

Concerning the first, plasticity is a local learning mechanism that regulates the growth of the neurons and their adaptation to stimuli, and it is also responsible for the memory mechanism~\cite{zilles1992neuronal, patten2015benefits}.

A biologically plausible plasticity model is the Hebbian model~\cite{brown1990hebbian, hebb2005organization}, which stems from the intuition that, if a synapse is often used, it should be strengthened. In other words, if the pre-synaptic neuron causes the activation of the post-synaptic one more frequently, the connection will become more relevant~\cite{caporale2008spike}.
Based on this idea, different Hebbian models have been developed~\cite{soltoggio2018born,coleman2012evolving,floreano2000evolutionary,yaman2021evolving}. Here, we consider in particular the so-called ABCD model, that has been proven effective in optimizing agents in several control tasks~\cite{mattiussi2007analog,niv2001evolution,soltoggio2007evolving,najarro2020meta,ferigo2022evolving}. 

Pruning, on the other hand, is a higher level process that is yet fundamental for cognitive development. For instance, it has been estimated that during puberty humans loose around $40\%$ of synapses without degrading memory or cognitive abilities~\cite{bourgeois_2010}.

In the AI community, pruning has recently attracted a lot of attention in the attempt to improve the performance and robustness, especially in the context of deep NNs~\cite{Hoefler, Chen_2021,frankle2018lottery,mocanu2018scalable, NEURIPS2019_e9874147,gale2019sparsity,liu2021sparse}.
Moreover, some other works have collected evidence on the fact that plasticity can effectively be seen as a form of implicit pruning, especially in cases where synaptic connections can saturate their values in such a way to obtain a quasi-binary mask on the weights~\cite{nadizar2021effects, nadizar2022merging,ferigo2022evolving,journe2022hebbian}. However, in these works, no explicit pruning mechanism has been employed.


\section{Methods}
\label{sec:methodology}
As introduced before, we aim to construct a network that can simulate the synaptogenesis process. In this section, we introduce the structure of this network and its behavior. Then, we briefly describe the optimization process and the classical control tasks used in our experimentation. We used these environments to measure the performance of the proposed SBNNs and to prove how they can change their structure during (and depending on) the task at hand.

\subsection{Hebbian learning}
\label{sec:HebbianUpdate}
Hebbian learning is a plasticity model that allows an NN to change its weights during the execution of a task. 
Importantly, this change is agnostic w.r.t. the reward for the task, because it is based only on the local knowledge of each synapse, in particular the activation of the pre-synaptic and post-synaptic neurons.
The ABCD model used in this work updates the weights after each forward pass of the network using the following rule:
\[
w_{i,j} = w_{i,j} + \eta(Aa_i + Ba_j + Ca_i a_j + D)
\]
where $a_i$ is the pre-synaptic activation value, $a_j$ is the post-synaptic value, and $w_{i,j}$ is the weight on the connection between the two neurons.
The $A$, $B$, $C$, and $D$ are parameters to optimize. 

\subsection{Pruning mechanism}
\label{sec:pruning}
The pruning mechanism aims to find a subset of an NN that performs as well as (or better than) the original network. This process is performed by removing connections based on a given strategy. 
In this work, we use the global magnitude pruning algorithm~\cite{Bishop1995NeuralNF,han2015pruning}, that simply consists in removing all the connections whose weights are smaller, in absolute value, than a threshold that is defined as the $pr$-th percentile, where $pr$ is the desired pruning rate (i.e., the percentage of connections to remove).
 
\subsection{Self-building Neural Network}
During synaptogenesis, neurons explore the extracellular space, assembling as many connections as possible with other neurons~\cite{petzoldt2014synaptogenesis,jin2005synaptogenesis,huttenlocher1997regional}.
Here, we aim to simulate this mechanism by allowing the possibility that any two neurons could directly connect.
 
Our model works as follows. We start from an NN composed of $I$ inputs, $H$ hidden nodes, and $O$ outputs. 
At the first episode of the task, the $I$ inputs are connected to all the $H$ hidden nodes and the $O$ outputs. In turn, the $H$ hidden nodes are fully connected with each other (excluding self-loops), and with all the $O$ outputs.
In total, the number of connections $C$, expressed as a function of $H$ (which is the only hyperparameter, as $I$ and $O$ depend on the task), is $C(H) = H^2 + H\times (I+O) + I\times O$.
Overall, the internal structure of the network resembles the one of the Boltzmann machines~\cite{ACKLEY1985147,hinton1986learning,mocanu2016topological,boltzIntro}, where the hidden nodes are fully connected. However, we also directly connect the inputs to the outputs, as the computational power of the hidden nodes in some cases may not be necessary.
 
We initialize the weights of all these connections to zero. In this way, we intend to simulate the initial condition where no connections between neurons exist. 
Then, \emph{within} each episode, the Hebbian procedure will update the weights based on the ABCD rule described in \Cref{sec:HebbianUpdate}. Note that, in our model, each connection in the network has its own ABCD rule with its corresponding parameters. In this way, the network can arrange itself based on the experience that the agent accumulates during the task. We use a Hebbian rule for each connection because, starting from a condition where all the weights are $0$, using a single Hebbian rule could lead all the weights to change in the same direction, which in turn would make learning ineffective.
 
The second step of synaptogenesis is the process that prunes the synapses, as described in \Cref{sec:pruning}. As we will describe later, differently from Hebbian learning, pruning occurs \emph{across} episodes. Note that, as soon as pruning starts, Hebbian learning is stopped.
\Cref{fig:ndrnn} summarizes the pruning procedure, showing the state of the initial network and its development.
Formally, we can analyze the network before and after pruning. 
In particular, before pruning, the hidden nodes are fully connected with each other and all the inputs are connected with all the hidden nodes. For this reason, it is not possible to define a fixed activation order. Hence, we maintain the overall order of activation: firstly, the inputs, then the hidden nodes, and then the outputs. However, for the hidden nodes, we randomly select the activation order.
 
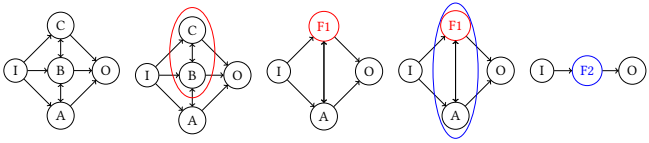
\begin{figure}[ht]
 \begin{tikzpicture}
 \node[scale=0.6] (S0) at (0,1) {\vt{\pic{ndrCyclepc={0}};}};
 \node[scale=0.6] (S0) at (1.75,1) {\vt{\pic{ndrCycleHLpc={0}};}};
 \node[scale=0.6] (S0) at (3.5,1) {\vt{\pic{ndrCycleFKpc={0}};}};
 \node[scale=0.6] (S0) at (5.25,1) {\vt{\pic{ndrCycleFHLpc={0}};}};
 \node[scale=0.6] (S0) at (7,1) {\vt{\pic{ndrCycleFKppc={0}};}};
 \end{tikzpicture}
 \caption{A minimal example of the simplification procedure described in \Cref{alg:removingCycles}. The base graph has two cycles: one composed of nodes $A$ and $B$, and one composed of nodes $B$ and $C$. The procedure identifies first the cycle ${B,C}$, circled in red, and replaces it with the red fake node $F_1$, which has incoming connections from $A$ and $I$ and outgoing connections to $A$ and $O$. Then, it finds the cycle ${F_1, A}$, circled in blue, and replaces it with the blue fake node $F_2$. Eventually, a directed acyclic graph is obtained, with a single fake hidden node $F_2$, which hides the cycle ${A,F_1}$, where $F_1$ is another fake node hiding the nodes $B$ and $C$.}
 \label{fig:graphSimplification}
\end{figure}

After pruning, the remaining connections define the network. Differently from before, in this phase we can define an activation order more easily because the pruning mechanism naturally resolves most cycles, especially if the ratio of connections removed is high enough. Hence, to find the activation order, we can perform a topological sort of the underlined graph $G(V, E)$, where $V$ is the set of nodes (i.e., the neurons) and $E$ is the set of connections.
If, during the topological sort, we find a cycle, we apply the following procedure. Indicating with $N_{c}$ the subset of $V$ that contains all the nodes in the cycle, first we remove all the nodes in $N_{c}$ from $V$ and replace them with a \emph{fake} node, $f$. Then, indicating with $E(N_{c})_{incoming} = \{(v, n)~\forall v\in V \setminus N_{c}~\land~ \forall n\in N_{c}\}$ the subset of $E$ composed of the connections that terminate in $N_{c}$ and that do not start from $N_{c}$, we add to $E$ the set of connections $\{(v,f)~\forall (v, v') \in E(N_{c})_{incoming}~\land~v' \in N_{c}\}$. We perform the same operation for the connections outgoing from $N_{c}$.
This procedure, that we apply iteratively and independently for every cycle found in the graph, results in a new graph $G'$ where the cycle $N_{c}$ is replaced by the fake node $f$. All the nodes connected to $N_{c}$ are now connected to $f$, and $f$ is connected to all the nodes reached from $N_{c}$.
We store the information that the node $f$ replaces the $N_{c}$ nodes in the $cycles\_history$ variable and then retry find a topological order. We repeat this procedure until all the cycles have been replaced, and a topological order can be defined. \Cref{alg:removingCycles} illustrates this simplification procedure. Note that the nodes in $N_{c}$ can also be fake nodes from a previous iteration of the procedure, as illustrated in \Cref{fig:graphSimplification}.

\begin{algorithm}[ht]
 \SetKwProg{Fn}{function}{\string:}{end}
 \SetKwFunction{removeCycle}{removeCycle}
 \SetKwFunction{findCycle}{findCycle}
 \SetKwFunction{removeNodes}{removeNodes}
 \SetKwFunction{incomingEdges}{incomingEdges}
 \SetKwFunction{outgoingEdges}{outgoingEdges}
 \SetKwFunction{addEdge}{addEdge}
 \SetKwFunction{removeEdge}{removeEdge}
 \SetKwFunction{addNode}{addNode}
 \SetKwFunction{addKeyValue}{addKeyValue}
 \Fn{$\removeCycle(G)$}{
 $N_c \gets \findCycle(G)$ \\
 $i \gets 0$ \\
 $cycleHistory \gets map()$
 \While{$N_c$ is not empty}{
 $E_{incoming} \gets G.\incomingEdges(N_{c})$ \\
 $E_{outgoing} \gets G.\outgoingEdges(N_{c})$ \\
 $G.\removeNodes(N_{c})$\\
 $G.\addNode(fakeNode_{i})$\\
 $cycleHistory.\addKeyValue(fakeNode_{i}, N_c)$\\
 \ForEach{$(v,v') \in E_{incoming}$}{
 $G.\removeEdge((v,v'))$\\
 $G.\addEdge((v,f))$\\
 }
 \ForEach{$(v,v') \in E_{outgoing}$}{
 $G.\removeEdge((v,v'))$\\
 $G.\addEdge((f,v'))$\\
 }
 $i \gets i +1$\\
 }
 \Return $G$
 }
 \caption{Pseudo code of the cycle removal procedure. The procedure replaces cycles with fake nodes while maintaining a history of the nodes that have been replaced.}
 \label{alg:removingCycles}
\end{algorithm}

After calculating the topological order of the network, we can follow that for the activation of the hidden nodes. If we find a fake node during the activation, we retrieve from $cycles\_history$ the set of $N_{c}$ nodes that compose the cycle, and proceed with a random activation order. If a node in $N_{c}$ is a fake node $f'$ covering the $N_{c'}$ cycle, we repeat the procedure solving the inner cycle $N_{c'}$, before continuing with the nodes in $N_{c}$. 

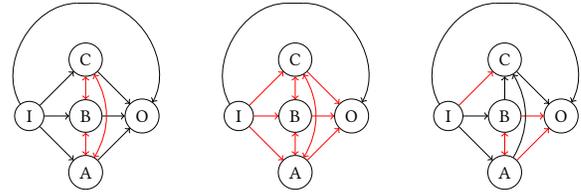
\begin{figure}[ht]
 \centering
 \begin{subfigure}[t]{0.30\columnwidth}
 \centering
 \begin{tikzpicture}
 \node[scale=0.75] (S0) at (1,1) {\vt{\pic{ndrcaseapc={}};}};
 \end{tikzpicture}
 \caption{Pruning removes inner connections between hidden nodes.}
 \label{fig:ndrnn_pruning_case_a}
 \end{subfigure}
 \hspace{0.01\columnwidth}
 \begin{subfigure}[t]{0.30\columnwidth}
 \centering
 \begin{tikzpicture}
  \node[scale=0.75] (S0) at (1,1) {\vt{\pic{ndrcasebpc={}};}};
 \end{tikzpicture}
 \caption{Pruning removes all connections from and to hidden nodes.}
 \label{fig:ndrnn_pruning_case_b}
 \end{subfigure}
 \hspace{0.01\columnwidth}
 \begin{subfigure}[t]{0.30\columnwidth}
 \centering
 \begin{tikzpicture}
  \node[scale=0.75] (S0) at (1,1) {\vt{\pic{ndrcasecpc={}};}};
 \end{tikzpicture}
 \caption{Pruning removes the connection between $I$ and $C$, creating two hidden layers.}
 \label{fig:ndrnn_pruning_case_c}
 \end{subfigure}
 \caption{Pruning can create different structures in the SBNN. In \Cref{fig:ndrnn_pruning_case_a}, an NN with one layer is created. In \Cref{fig:ndrnn_pruning_case_b}, pruning cuts all the connections to the inner nodes, reducing the NN only to the input-output connections. Finally, in \Cref{fig:ndrnn_pruning_case_c}, pruning results in the creation of two hidden layers.}
 \label{fig:fig:ndrnn_pruning}
\end{figure}

Thanks to this process, the network after pruning can have a different structure. We identify three base structures, which can be described as follows.
In the first case, the inputs are connected to all the hidden nodes that in turn are connected to the outputs. This creates an NN with a single hidden layer, see \Cref{fig:ndrnn_pruning_case_a}. 
In the second case, the pruning process cuts all the connections between the input and hidden nodes. Hence, the inputs are directly connected to the output nodes, creating a zero-layer NN, see \Cref{fig:ndrnn_pruning_case_b}.
In the third case, the pruning process removes all the connections between the inputs and a subset of the hidden nodes, but the hidden nodes remain connected, creating an NN with more than one layer. For example, given the hidden nodes $A$, $B$, and $C$, if the inputs are connected only with $A$ and $B$, but not with $C$, which in turn remains connected with $A$ and $B$, the resulting NN will have two hidden layers: the first one, composed of neurons $A$ and $B$; the second one, composed only of neuron $C$, see \Cref{fig:ndrnn_pruning_case_c}.

It is worth noticing that, starting from these three base structures, we can derive more complex structures. For example, each node in the third case can be a \emph{fake} node, hence ``hiding'' a set of nodes.

\begin{figure}[ht]
 \centering
 \begin{tikzpicture}
 \begin{axis}[
 axis lines = left,
 xlabel = {Hidden nodes},
 ylabel = {Parameters},
 legend columns=3,
 width=0.5\columnwidth,
 height=0.5\columnwidth,
 legend entries={FFNN, Hebbian FFNN, SBNN},
 legend to name=legendComparison
 ]
 \addplot [
 domain=0:100, 
 samples=100, 
 color=red,
 ]
 {(x/2)^2 + (x/2)};
 \addplot [
 domain=0:100, 
 samples=100, 
 color=green,
 ]
 {4*((x/2)^2 +(x/2))};
 \addplot [
 domain=0:100, 
 samples=100, 
 color=blue,
 ]
 {4*(x^2 + 2*x + 2)};
 \end{axis}
 \end{tikzpicture}
 \pgfplotslegendfromname{legendComparison}
 \caption{Number of parameters to optimize with respect to the number of hidden nodes. For the (Hebbian) FFNN, we consider an NN with two hidden layers with the same number of neurons indicated on the x-axis. In all cases, a single input and a single output are considered. The FFNN with Hebbian learning uses a different ABCD rule for each connection.}
 \label{fig:weightsComparison}
 \end{figure}
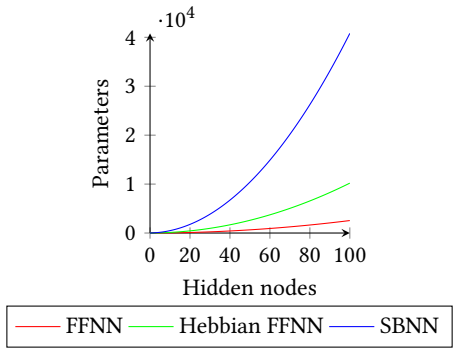

 
While in principle promising, this model is not free from weak points. 
First of all, the number of weights in the SBNN increases quadratically with the number of hidden nodes, as each hidden node is fully connected with all the other hidden nodes. Moreover, each connection is associated with an ABCD rule with its corresponding $4$ parameters. Therefore, the total number of parameters to optimize is orders of magnitude greater with respect to a Feed Forward Neural Network (FFNN) with the same number of hidden nodes. \Cref{fig:weightsComparison} shows a comparison in terms of number of parameters to optimize between the proposed SBNN, and two other NN-based models, namely an FFNN with and without Hebbian learning.
 
Secondly, before pruning the hidden nodes compose a single, fully connected subnetwork on which the activation order can influence the network's output. For example, if we consider a fixed activation order, the first hidden node will receive as inputs the values of the inputs at the current timestep, while, for the other hidden nodes, the values received as inputs will be the ones from the previous activation. On the other hand, as discussed above, after pruning we can find the topological order for network by visiting the underlying graph. Still, in the presence of a subset of hidden nodes that are linked together, we cannot determine a unique activation order.
 
\subsection{OpenAI tasks}
To measure the performance of the proposed SBNN, we use three classical control tasks from OpenAI~\cite{brockman2016openai}, namely \emph{Cart Pole}, \emph{Mountain Car}, and \emph{Lunar Lander}. 
 
In Cart Pole, the agent has to move a cart to maintain a pole in equilibrium. 
The agent can push the cart in both directions (move left/right), and it receives a positive reward for each timestep in which the pole is in equilibrium.
The episode ends after $500$ timesteps, or if the angle of the pole is outside the range $\pm 12\degree$.
 
In Mountain Car, the agent has to drive a car from a valley to the top of a mountain. The agent has to build momentum to increase its velocity, thanks to another hill positioned before it. 
The agent can perform three actions (accelerate left, accelerate right, or do not accelerate), and receives a negative reward at each timestep until it reaches the top of the mountain.
The episode ends if the car reaches the top of the hill, or after $200$ timesteps.

Finally, in Lunar Lander, the agent has to land a spaceship. The agent has to reduce the terminal velocity of the spaceship while compensating for the lateral wind. 
Hence, the agent can perform four actions: two to control the lateral (left/right) engines, one to activate the main engine, and one that does not perform any action.
The agent increases the received reward if it lands in the designated area at a lower speed and does not tilt.
The episode ends if the spaceship lands, or if its $x$ position is greater than $1$.
 
The three tasks above are solved if the average reward over $100$ episodes is greater than a predefined threshold, which is $475$ for Cart Pole, $-110$ for Mountain Car, and $200$ for Lunar Lander.

\begin{figure}[ht]
 \begin{tikzpicture}
 \node[inner sep=0pt] (LS) at (0,0) {};
 \node[inner sep=0pt] (LE) at (0.9\columnwidth,0) {};
 \draw[|->] (LS) to node[below] {Episodes} (LE) ;
 \node[scale=0.5] (S0) at (1,1) {\vt{\pic{ndrStart={0 0}};}};
 \node[scale=0.5] (S1) at (3,1) {\vt{\pic{ndr1step={0 0}};}};
 \node[scale=0.5] (S2) at (6,1) {\vt{\pic{ndrpp={0 0}};}};
 \end{tikzpicture}
 \caption{Scheme of the SBNN over the episodes. Initially, we set all the connections to $0$ (red); then during the task Hebbian plasticity changes the weights, leading to the second NN, where different thickness indicates different weights. At a certain time, the pruning mechanism cuts the weakest connections, resulting in the final structure of the NN.}
 \label{fig:ndrnn}
\end{figure}
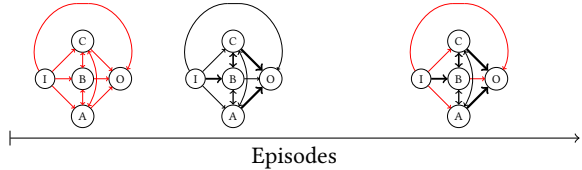

\section{Results}
\label{sec:results}
In this section, we will analyze the performance and behavior of the SBNN in comparison with a FFNN for which we apply the same pruning mechanism used in the SBNN, but before the fitness evaluation. Note that, in our implementation, we use as activation function the $tanh$ function, both on the hidden nodes and on the input/output nodes, for both the SBNN and the FFNN.
Concerning the fitness evaluation, we measure the performance of the agent as the average reward over $100$ episodes seen during training. 
For the FFNN, as the pruning process happens before the first episode, in this case the fitness by construction is measured after pruning. On the contrary, as for the SBNN the pruning process happens during the life of the agent (i.e., across episodes), in this case the fitness contains two components, one before and one after pruning (which are then averaged).
We divide our experiments into three parts, to answer three different research questions:
\begin{enumerate}[leftmargin=22pt]
 \item[\textbf{RQ1}] What is the performance of the SBNN? What are the main hyperparameters of this model that affect the performance?
 \item[\textbf{RQ2}] Is there any structural difference between the networks produced by the SBNN and an FFNN?
 \item[\textbf{RQ3}] Are SBNNs able to generalize over the different tasks?
\end{enumerate}
 
To answer these questions, we perform a campaign of simulations varying the three main hyperparameters of the SBNN: the number of hidden nodes $hn$, the pruning rate $pr$, and the pruning time $pt$, the latter indicating when pruning is applied. 
To calculate the pruning time, we consider the number of episodes in the task, i.e., a pruning time of $10$ means that pruning happens after the agent completes the $10$-th episode.
 
For each combination of these parameters, we perform $30$ independent evolutionary processes. To optimize the parameters of the network (i.e., the weights for the FFNN, or the parameters of the ABCD rules for the SBNN), we use the well-known Covariance Matrix Adaptation Evolution Strategies (CMA-ES)~\cite{cmaes,loshchilov2016cma,hansen1996adapting,fontaine2020covariance}.
We stop the evolution after the generation of a fixed number of $2000$ individuals for Lunar Lander and Cart Pole, and $4000$ for Mountain Car. In all cases, we set $\lambda = 4+ \lfloor 3* ln(\mathbf{|p|}) \rfloor$ and $\mu = \frac{\lambda}{2}$, where $\mathbf{p}$ is the vector of parameters to optimize. \Cref{tab:params} summarizes the configurations tested.
Note that we omit the results obtained on the Cart Pole task as there were no significant differences between the FFNN and the SBNN: in fact, all the individuals using both models solved the task, that is comparatively simpler than the other two.
We make our code publicly available at \url{https://github.com/ndr09/SBM}.

\begin{table}[ht!]
 \centering
 \begin{tabular}{
 l
 S[table-format=4.0]
 S[table-format=4.0]
 S[table-format=4.0]
 }
 \toprule
 Parameter & {Cart Pole}& {Mountain Car} & {Lunar Lander}\\
 \midrule
 Fitness evaluation & 2000 & 4000 & 2000 \\
 Hidden nodes & {3, 4} & {3, 4} & {5, 6, 7, 8, 9} \\
 Pruning time & {5, 10} & {1, 5, 10} & {1, 5, 10, 15, 20} \\
 Pruning rate & {40, 60} & {40, 60} & {20, 40, 60, 80} \\
 \bottomrule
 \end{tabular}
 \caption{
 Parameter configuration used for each task considered in the RQ1 experiments.
 For RQ2 and RQ3, we use a representative subset of these configurations.
 }
 \label{tab:params}
\end{table}

\subsection{RQ1: Performance}
Concerning the performance of the SBNN, we aim to evaluate how it compares with that of an FFNN. Since the SBNN has more connections than the FFNN given the same number of hidden nodes, we make two comparisons: one comparing the results of the two models given the same number of hidden nodes, and one given the same total number of connections after pruning.
In the following, we start with the Mountain Car task and then move to describe the results for the Lunar Lander one.
 
\begin{figure}[ht]
 \centering
 \begin{tikzpicture}
 \begin{groupplot}[
 boxplot,
 boxplot/draw direction=y,
 group style={
 group size=2 by 3,
 horizontal sep=1mm,
 vertical sep=1.5mm,
 xticklabels at=edge bottom,
 yticklabels at=edge left
 },
 ymin=-150,
 ymax=-100,
 width=0.5\columnwidth,
 height=0.5\columnwidth,
 legend cell align={center},
 xmajorticks=false
 ]
 \nextgroupplot[
 align=center,
 title={pruning rate $40\%$},
 ylabel = {Hidden nodes $3$ \\ Reward},
 legend columns=4,
 legend entries={PT 10, PT 5, PT 1, FFNN},
 legend style={draw=none},
 legend to name=mctrain
 ]
 \addlegendimage{area legend,color=cola7,fill}
 \addlegendimage{area legend,color=cola2,fill}
 \addlegendimage{area legend,color=cola1,fill}
 \addlegendimage{area legend,color=cola3,pattern={north east lines},pattern color=cola3}
 \addplot[black,fill=cola7] table[y=10_40_3] {data/mc_sbm_train.txt};
 \addplot[black,fill=cola2] table[y=5_40_3] {data/mc_sbm_train.txt};
 \addplot[black,fill=cola1] table[y=1_40_3] {data/mc_sbm_train.txt};
 \addplot[black,fill=cola3,pattern={north east lines},pattern color=cola3] table[y=40_3] {data/mc_nn_train.txt};
 \draw [dashed, red] (\pgfkeysvalueof{/pgfplots/xmin}, -110) -- (\pgfkeysvalueof{/pgfplots/xmax}, -110);
 \nextgroupplot[
 align=center,
 title={pruning rate $60\%$}
 ]
 \addplot[black,fill=cola7] table[y=10_60_3] {data/mc_sbm_train.txt};
 \addplot[black,fill=cola2] table[y=5_60_3] {data/mc_sbm_train.txt};
 \addplot[black,fill=cola1] table[y=1_60_3] {data/mc_sbm_train.txt};
 \addplot[black,fill=cola3,pattern={north east lines},pattern color=cola3] table[y=60_3] {data/mc_nn_train.txt};
 \draw [dashed, red] (\pgfkeysvalueof{/pgfplots/xmin}, -110) -- (\pgfkeysvalueof{/pgfplots/xmax}, -110);
 \nextgroupplot[
 align=center,
 ylabel = {Hidden nodes $4$ \\ Reward},
 ]
 \addplot[black,fill=cola7] table[y=10_40_4] {data/mc_sbm_train.txt};
 \addplot[black,fill=cola2] table[y=5_40_4] {data/mc_sbm_train.txt};
 \addplot[black,fill=cola1] table[y=1_40_4] {data/mc_sbm_train.txt};
 \addplot[black,fill=cola3,pattern={north east lines},pattern color=cola3] table[y=40_4] {data/mc_nn_train.txt};
 \draw [dashed, red] (\pgfkeysvalueof{/pgfplots/xmin}, -110) -- (\pgfkeysvalueof{/pgfplots/xmax}, -110);
 \nextgroupplot[
 align=center
 ]
 \addplot[black,fill=cola7] table[y=10_60_4] {data/mc_sbm_train.txt};
 \addplot[black,fill=cola2] table[y=5_60_4] {data/mc_sbm_train.txt};
 \addplot[black,fill=cola1] table[y=1_60_4] {data/mc_sbm_train.txt};
 \addplot[black,fill=cola3,pattern={north east lines},pattern color=cola3] table[y=60_4] {data/mc_nn_train.txt};
 \draw [dashed, red] (\pgfkeysvalueof{/pgfplots/xmin}, -110) -- (\pgfkeysvalueof{/pgfplots/xmax}, -110);
 \end{groupplot}
 \end{tikzpicture}\\
 \pgfplotslegendfromname{mctrain}
 \caption{Results on the Mountain Car environment. The y-axis shows the average reward obtained during training from the best agent found in each of $30$ runs. The dashed line indicates the solving threshold for the environment. For the SBNN, the results indicate a clear trend where the performance increases while $pt$ decreases.}
 \label{fig:mc_train}
 \end{figure}
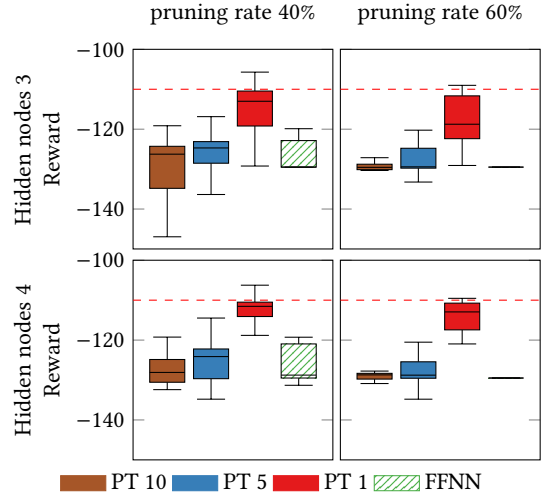

\Cref{fig:mc_train} shows the results for the Mountain Car environment. The upper and lower row relate, respectively, to an FFNN with one layer with $3$ or $4$ hidden nodes. The left and right column present, respectively, the results with a pruning rate of $40\%$ and $60\%$. In each subfigure, we plot the average results of the best individual over $30$ independent runs. The first three boxplots indicate the results of the SBNN with different pruning times, namely $10$, $5$, and $1$, respectively from left to right. The last boxplot shows the baseline results of the FFNN. 
In Mountain Car, the results indicate that the SBNN reaches similar or better performance with respect to an FFNN with the same pruning rate.
 
Interestingly, we observe a clear trend on the pruning time, i.e., the performance increases when decreasing the value of $pt$, regardless of the pruning rate and the number of hidden nodes. 
Hence, we can conclude that, after the first episode, the agent has already received enough experience (information) to build the network. 

To understand the effect of pruning on the performance, we make an additional analysis, by comparing the average reward before and after pruning. For instance, considering $pt=10$, we measure separately the average reward until the $10$-th episode, i.e., before pruning, and the average reward after the $10$-th episode, i.e., the post-pruning one.
Based on this procedure, we observe that, thanks to pruning, the performance receives a $7-13\%$ boost on the Mountain Car task.
 
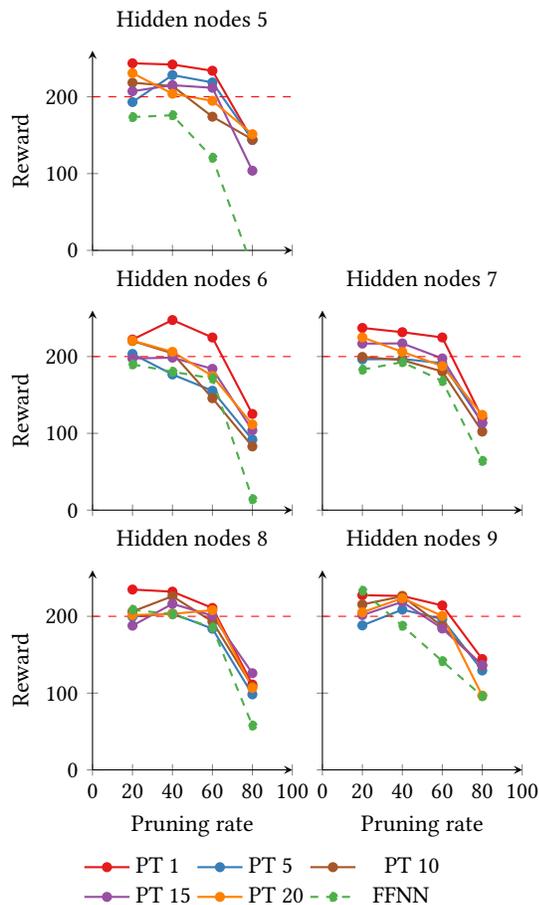
\begin{figure}[ht]
 \centering
 \begin{tikzpicture}
 \begin{groupplot}[
 axis lines = left,
 group style={
 group size=2 by 3,
 horizontal sep=4mm,
 vertical sep=8mm,
 xticklabels at=edge bottom,
 yticklabels at=edge left
 },
 xmin=0,
 xmax=100,
 ymin=0,
 ymax=260,
 width=0.5\columnwidth,
 height=0.5\columnwidth
 ]
  \addlegendimage{color=cola1,fill}
 \addlegendimage{color=cola2,fill}
 \addlegendimage{color=cola7,fill}
 \addlegendimage{color=cola4,fill}
 \addlegendimage{color=cola5,fill}
 \addlegendimage{color=cola3,fill}
 \nextgroupplot[
 align=center,
 title={Hidden nodes $5$},
 legend columns=3,
 legend entries={PT 1\hphantom{5}, PT 5\hphantom{5}, PT 10, PT 15, PT 20, \hphantom{6}FFNN\hphantom{555}},
 legend style={draw=none},
 legend to name=legendComparisonLL,
 ylabel = {Reward}
 ]
 \linesimple{data/ll_sbm_train_asd_nf.txt}{pr}{5_1_m}{cola1};
 \linesimple{data/ll_sbm_train_asd.txt}{pr}{5_5_m}{cola2};
 \linesimple{data/ll_sbm_train_asd.txt}{pr}{5_10_m}{cola7};
 \linesimple{data/ll_sbm_train_asd.txt}{pr}{5_15_m}{cola4};
 \linesimple{data/ll_sbm_train_asd.txt}{pr}{5_20_m}{cola5};
 
 \linesimpledash{data/ll_nn_train_asd.txt}{pr}{5_m}{cola3};
 \draw [dashed, red] (\pgfkeysvalueof{/pgfplots/xmin}, 200) -- (\pgfkeysvalueof{/pgfplots/xmax}, 200);
 \nextgroupplot[
 group/empty plot
 ]
 \nextgroupplot[
 align=center,
 title={Hidden nodes $6$},
 ylabel = {Reward}
 ]
 \linesimple{data/ll_sbm_train_asd_nf.txt}{pr}{6_1_m}{cola1};
 \linesimple{data/ll_sbm_train_asd.txt}{pr}{6_5_m}{cola2};
 \linesimple{data/ll_sbm_train_asd.txt}{pr}{6_10_m}{cola7};
 \linesimple{data/ll_sbm_train_asd.txt}{pr}{6_15_m}{cola4};
 \linesimple{data/ll_sbm_train_asd.txt}{pr}{6_20_m}{cola5};
 
 \linesimpledash{data/ll_nn_train_asd.txt}{pr}{6_m}{cola3};
 \draw [dashed, red] (\pgfkeysvalueof{/pgfplots/xmin}, 200) -- (\pgfkeysvalueof{/pgfplots/xmax}, 200);
 \nextgroupplot[
 align=center,
 title={Hidden nodes $7$}
 ]
 \linesimple{data/ll_sbm_train_asd_nf.txt}{pr}{7_1_m}{cola1};
 \linesimple{data/ll_sbm_train_asd.txt}{pr}{7_5_m}{cola2};
 \linesimple{data/ll_sbm_train_asd.txt}{pr}{7_10_m}{cola7};
 \linesimple{data/ll_sbm_train_asd.txt}{pr}{7_15_m}{cola4};
 \linesimple{data/ll_sbm_train_asd.txt}{pr}{7_20_m}{cola5};
 
 \linesimpledash{data/ll_nn_train_asd.txt}{pr}{7_m}{cola3};
 \draw [dashed, red] (\pgfkeysvalueof{/pgfplots/xmin}, 200) -- (\pgfkeysvalueof{/pgfplots/xmax}, 200);
 \nextgroupplot[
 align=center,
 title={Hidden nodes $8$},
 ylabel = {Reward},
 xlabel = {Pruning rate}
 ]
 \linesimple{data/ll_sbm_train_asd_nf.txt}{pr}{8_1_m}{cola1};
 \linesimple{data/ll_sbm_train_asd.txt}{pr}{8_5_m}{cola2};
 \linesimple{data/ll_sbm_train_asd.txt}{pr}{8_10_m}{cola7};
 \linesimple{data/ll_sbm_train_asd.txt}{pr}{8_15_m}{cola4};
 \linesimple{data/ll_sbm_train_asd.txt}{pr}{8_20_m}{cola5};
 
 \linesimpledash{data/ll_nn_train_asd.txt}{pr}{8_m}{cola3};
 \draw [dashed, red] (\pgfkeysvalueof{/pgfplots/xmin}, 200) -- (\pgfkeysvalueof{/pgfplots/xmax}, 200);
 \nextgroupplot[
 align=center,
 title={Hidden nodes $9$},
 xlabel = {Pruning rate}
 ]
 \linesimple{data/ll_sbm_train_asd_nf.txt}{pr}{9_1_m}{cola1};
 \linesimple{data/ll_sbm_train_asd.txt}{pr}{9_5_m}{cola2};
 \linesimple{data/ll_sbm_train_asd.txt}{pr}{9_10_m}{cola7};
 \linesimple{data/ll_sbm_train_asd.txt}{pr}{9_15_m}{cola4};
 \linesimple{data/ll_sbm_train_asd.txt}{pr}{9_20_m}{cola5};
 
 \linesimpledash{data/ll_nn_train_asd.txt}{pr}{9_m}{cola3};
 \draw [dashed, red] (\pgfkeysvalueof{/pgfplots/xmin}, 200) -- (\pgfkeysvalueof{/pgfplots/xmax}, 200);
 \end{groupplot}
 \end{tikzpicture}
 \pgfplotslegendfromname{legendComparisonLL}
 \caption{Median reward on the Lunar Lander task for different numbers of hidden nodes. The x-axis indicates the pruning rate, while the different lines refer to different values of pruning time. The red dashed line indicates the solving threshold. The results show that while increasing the pruning rate, the performance drop for the SBNN is lower than for the FFNN.}
 \label{fig:ll_train}
 \end{figure}

\Cref{fig:ll_train} presents the results for the Lunar Lander environment. For this environment, we consider hidden nodes ranging from $5$ to $9$, because of the greater complexity of the task.
Here, the results are shown differently from \Cref{fig:mc_train}: in particular, we plot the median rewards for the best individuals of each evolutionary run, varying the pruning rate while keeping the pruning time and the number of hidden nodes fixed.
In this way, we highlight two points. The first one is that the performances of the SBNNs are in most cases equal to or better than the ones obtained by the FFNN for the same pruning rate. The second point is that, while increasing the pruning rate, the drop in performance for the SBNN is lower than what observed with the FFNN baseline.
We can also observe that this trend is maintained when comparing solutions with a comparable number of connections. 
For example, the SBNN with $5$ hidden nodes and the FFNN with $9$ hidden nodes have a similar number of connections (respectively, $117$ and $108$).
Moreover, we can see the same trend observed in Mountain Car, i.e., that the best results are achieved with $pt = 1$. Hence, also in this task it appears that a single episode contains enough information to learn about the prunable connections. These observations suggest that, at least in the tested tasks, the SBNN is effectively capable to exploit the network structure (and the information therein) better than the FFNN.

Also in this case, we perform the same analysis of before, dividing the results before and after pruning and observing the average results. As for the case of Mountain Car, also for Lunar Lander we can see an increase in performance after pruning, in this case between $6\%$ and $100\%$. This improvement indicates, once again, the importance of pruning and its complementary effect with respect to Hebbian learning.

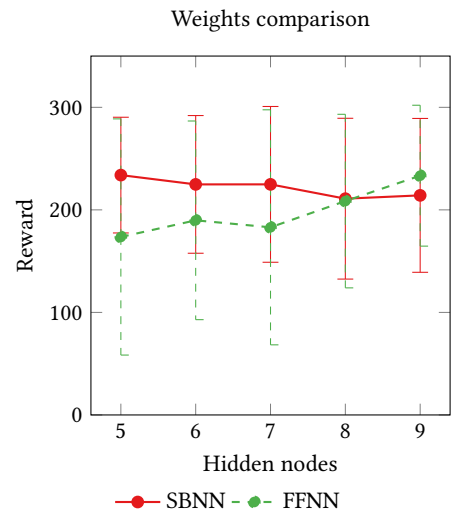
\begin{figure}[ht]
 \centering
 \begin{tikzpicture}
 \begin{groupplot}[
 group style={
 group size=1 by 1,
 horizontal sep=1mm,
 vertical sep=1.5mm,
 xticklabels at=edge bottom,
 yticklabels at=edge left
 },
 ymin=0,
 ymax=350,
 width=0.75\columnwidth,
 height=0.75\columnwidth,
 legend cell align={center},
 xtick distance=1
 ]
 
 \addlegendimage{color=cola1, fill}
 \addlegendimage{color=cola3, fill}
 \nextgroupplot[
 align=center,
 title={Weights comparison},
 ylabel = {Reward},
 xlabel = {Hidden nodes},
 legend columns=2,
 legend entries={SBNN, FFNN},
 legend style={draw=none},
 legend to name=swc
 ]
 \linevb{data/wc_com.txt}{hn}{sbm}{sbm_s}{cola1};
 \linesdashvb{data/wc_com.txt}{hn}{nn}{nn_s}{cola3};
 \end{groupplot}
 \end{tikzpicture}\\
 \pgfplotslegendfromname{swc}
 \caption{Average reward on the Lunar Lander task for a SBNN with a pruning rate of $60\%$ and an FFNN with a pruning rate of $20\%$. The x-axis indicates the number of hidden nodes (note that in this setting the two models have a similar number of total connections after the pruning process). The results show that the SBNN reaches similar or slightly better performances with respect to the FFNN. For $9$ hidden nodes, the FFNN seems to perform slightly better, but the difference is not statistically significant (Wilcoxon rank-sum test, $\alpha = 0.01$).}
 \label{fig:sameWeightsCmp}
 \end{figure}

Finally, we compare the results of the networks based on total number of connections after pruning. In \Cref{fig:sameWeightsCmp}, we compare the FFNN with $pr=20\%$ with the SBNN with $pr=60\%$ (with these values, in fact, the FFNN and the SBNN have a similar number of connections after pruning). We can observe that the SBNN reaches almost always a comparable or slightly better performance with respect to the FFNN. However, in the case where the FFNN performs better than the SBNN (i.e., with $9$ nodes), the difference is not statistically significant (Wilcoxon rank-sum test, $\alpha = 0.01$). 

\subsection{RQ2: Difference between SBNN and FFNN}
In this section, we want to analyze the structural difference between the network found with the SBNN and the FFNN. 
For this reason, we characterize the networks based on the number of \emph{working connections}, i.e., the connections that link inputs to outputs after pruning. To calculate these connections, we remove the synapses that lead to sink or come from source nodes. While a sink is a node with only incoming connections that is not an output node, a source is a node with only outgoing connections that is not an input node.

\begin{figure}
 \centering
 \begin{tikzpicture}
 \begin{groupplot}[
 width=0.55\linewidth,
 height=0.45\linewidth,
 grid=both,
 grid style={line width=.1pt, draw=gray!10},
 major grid style={line width=.2pt,draw=gray!50},
 title style={anchor=north, yshift=2ex},
 group style={
 group size=2 by 2,
 horizontal sep=3.5mm,
 vertical sep=2mm,
 xticklabels at=edge bottom,
 yticklabels at=edge left
 },
 ymax=1.1,
 ymin=0,
 xmax=100,
 xmin=0,
 ytick={0,0.5, 1}
 ]
 \addlegendimage{area legend,color=cola1,fill}
 \addlegendimage{area legend,color=cola2,fill}
 \addlegendimage{area legend,color=cola7,fill}
 \addlegendimage{area legend,color=cola4,fill}
 \nextgroupplot[ legend columns=4,
 legend entries={PT 1, PT 5, PT 10, FFNN},
 legend style={draw=none},
 legend to name=wchmc,title={Hidden nodes $5$}, ylabel={$pr=20\%$}]
 \linesimple{data/mc_sbm_wch.txt}{c}{1_40_3}{cola1};
 \linesimple{data/mc_sbm_wch.txt}{c}{5_40_3}{cola2};
 \linesimple{data/mc_sbm_wch.txt}{c}{10_40_3}{cola7};
 \linesimpledash{data/mc_nn_wch.txt}{c}{3_40}{cola3};
 \nextgroupplot[ title={Hidden nodes $4$}]
 \linesimple{data/mc_sbm_wch.txt}{c}{1_40_4}{cola1};
 \linesimple{data/mc_sbm_wch.txt}{c}{5_40_4}{cola2};
 \linesimple{data/mc_sbm_wch.txt}{c}{10_40_4}{cola7};
 \linesimpledash{data/mc_nn_wch.txt}{c}{4_40}{cola3};
 \nextgroupplot[ ylabel={$pr=60\%$}]
 \linesimple{data/mc_sbm_wch.txt}{c}{1_60_3}{cola1};
 \linesimple{data/mc_sbm_wch.txt}{c}{5_60_3}{cola2};
 \linesimple{data/mc_sbm_wch.txt}{c}{10_60_3}{cola7};
 \linesimpledash{data/mc_nn_wch.txt}{c}{4_60}{cola3};
 \nextgroupplot[ ]
 \linesimple{data/mc_sbm_wch.txt}{c}{1_60_4}{cola1};
 \linesimple{data/mc_sbm_wch.txt}{c}{10_60_4}{cola2};
 \linesimple{data/mc_sbm_wch.txt}{c}{10_60_4}{cola7};
 \linesimpledash{data/mc_nn_wch.txt}{c}{4_60}{cola3};
 \end{groupplot}
 \end{tikzpicture}
 \pgfplotslegendfromname{wchmc}
 \caption{Distribution of the number of connections after pruning the network on the Mountain Car task. On the x-axis, we show the percentage of working connections, which are connections that do not lead to a sink node or are outgoing from a source node, grouped every $10\%$. On the y-axis, we show how many networks (out of $30$, one per run) have that percentage of working connections. The results show that, for a pruning rate higher than $20\%$, FFNN tends to use all the available connections, while SBNN has two peaks: one around $10-20\%$, and one at $80-90\%$.}
 \label{fig:weights-divergence-mc}
 \vspace{-0.5cm}
\end{figure}
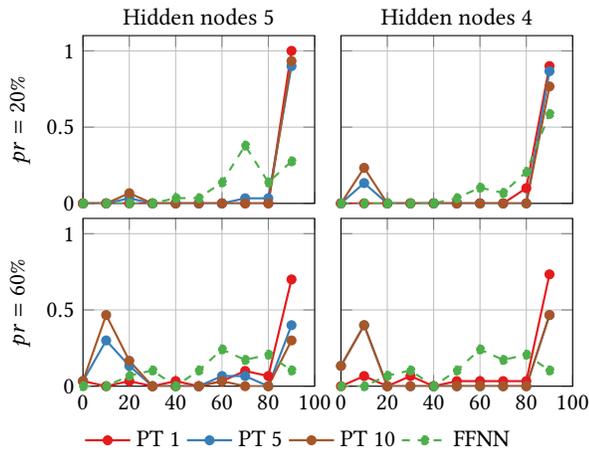

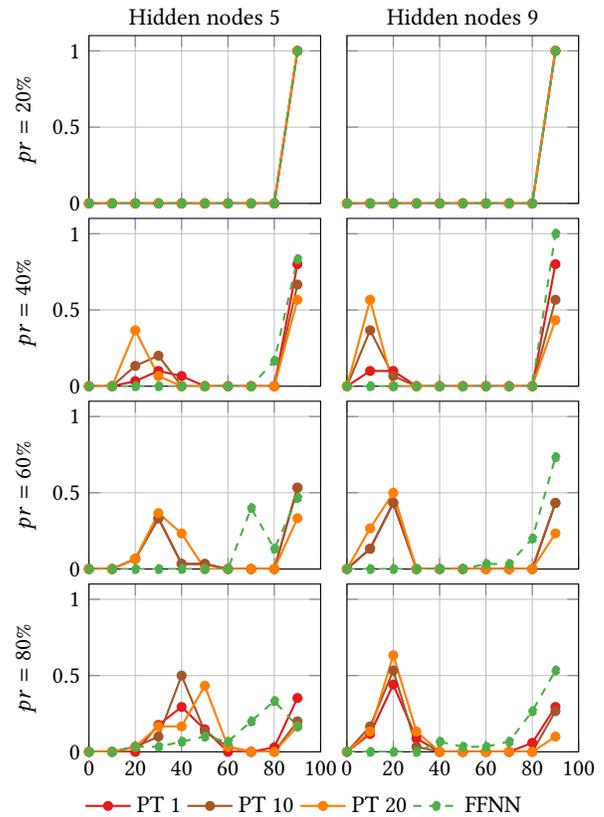
\begin{figure}
 \centering
 \begin{tikzpicture}
 \begin{groupplot}[
 width=0.55\linewidth,
 height=0.45\linewidth,
 grid=both,
 grid style={line width=.1pt, draw=gray!10},
 major grid style={line width=.2pt,draw=gray!50},
 title style={anchor=north, yshift=2ex},
 group style={
 group size=2 by 4,
 horizontal sep=3.5mm,
 vertical sep=2mm,
 xticklabels at=edge bottom,
 yticklabels at=edge left
 },
 ymax=1.1,
 ymin=0,
 xmax=100,
 xmin=0,
 ytick={0,0.5, 1}
 ]
 \addlegendimage{area legend,color=cola1,fill}
 \addlegendimage{area legend,color=cola2,fill}
 \addlegendimage{area legend,color=cola4,fill}
 \addlegendimage{area legend,color=cola3,fill}
 \nextgroupplot[ legend columns=4,
 legend entries={PT 1, PT 10, PT 20, FFNN},
 legend style={draw=none},
 legend to name=wchll,title={Hidden nodes $5$}, ylabel={$pr=20\%$}]
 \linesimple{data/ll_sbm_wch.txt}{c}{1_20_5}{cola1};
 \linesimple{data/ll_sbm_wch.txt}{c}{10_20_5}{cola7};
 \linesimple{data/ll_sbm_wch.txt}{c}{20_20_5}{cola5};
 \linesimpledash{data/ll_nn_wch.txt}{c}{5_20}{cola3};
 \nextgroupplot[ title={Hidden nodes $9$}]
 \linesimple{data/ll_sbm_wch.txt}{c}{1_20_9}{cola1};
 \linesimple{data/ll_sbm_wch.txt}{c}{10_20_9}{cola7};
 \linesimple{data/ll_sbm_wch.txt}{c}{20_20_9}{cola5};
 \linesimpledash{data/ll_nn_wch.txt}{c}{9_20}{cola3};
 \nextgroupplot[ ylabel={$pr=40\%$}]
 \linesimple{data/ll_sbm_wch.txt}{c}{1_40_5}{cola1};
 \linesimple{data/ll_sbm_wch.txt}{c}{10_40_5}{cola7};
 \linesimple{data/ll_sbm_wch.txt}{c}{20_40_5}{cola5};
 \linesimpledash{data/ll_nn_wch.txt}{c}{5_40}{cola3};
 \nextgroupplot[ ]
 \linesimple{data/ll_sbm_wch.txt}{c}{1_40_9}{cola1};
 \linesimple{data/ll_sbm_wch.txt}{c}{10_40_9}{cola7};
 \linesimple{data/ll_sbm_wch.txt}{c}{20_40_9}{cola5};
 \linesimpledash{data/ll_nn_wch.txt}{c}{9_40}{cola3};
 \nextgroupplot[ ylabel={$pr=60\%$}]
 \linesimple{data/ll_sbm_wch.txt}{c}{1_60_5}{cola1};
 \linesimple{data/ll_sbm_wch.txt}{c}{10_60_5}{cola7};
 \linesimple{data/ll_sbm_wch.txt}{c}{20_60_5}{cola5};
 \linesimpledash{data/ll_nn_wch.txt}{c}{5_60}{cola3};
 \nextgroupplot[ ]
 \linesimple{data/ll_sbm_wch.txt}{c}{1_60_9}{cola1};
 \linesimple{data/ll_sbm_wch.txt}{c}{10_60_9}{cola7};
 \linesimple{data/ll_sbm_wch.txt}{c}{20_60_9}{cola5};
 \linesimpledash{data/ll_nn_wch.txt}{c}{9_60}{cola3};
 \nextgroupplot[ ylabel={$pr=80\%$}]
 \linesimple{data/ll_sbm_wch.txt}{c}{1_80_5}{cola1};
 \linesimple{data/ll_sbm_wch.txt}{c}{10_80_5}{cola7};
 \linesimple{data/ll_sbm_wch.txt}{c}{20_80_5}{cola5};
 \linesimpledash{data/ll_nn_wch.txt}{c}{5_80}{cola3};
 \nextgroupplot[ ]
 \linesimple{data/ll_sbm_wch.txt}{c}{1_80_9}{cola1};
 \linesimple{data/ll_sbm_wch.txt}{c}{10_80_9}{cola7};
 \linesimple{data/ll_sbm_wch.txt}{c}{20_80_9}{cola5};
 \linesimpledash{data/ll_nn_wch.txt}{c}{9_80}{cola3};
 \end{groupplot}
 \end{tikzpicture}
 \pgfplotslegendfromname{wchll}
 \caption{Distribution of the number of connections after pruning the network on the Lunar Lander task. On the x-axis, we show the percentage of working connections, which are connections that do not lead to a sink node or are outgoing from a source node, grouped every $10\%$. On the y-axis, we show how many networks (out of $30$, one per run) have that percentage of working connections. The results show that, for a pruning rate higher than $20\%$, FFNN tends to use all the available connections, while SBNN has two peaks: one around $20-40\%$, and one at $80-90\%$.}
 \label{fig:weights-divergence-ll}
 \vspace{-0.5cm}
\end{figure}

\Cref{fig:weights-divergence-mc} and \Cref{fig:weights-divergence-ll} show the distribution of working connections for a representative subset of the configurations presented in the previous section, considering the best solution (one per each evolutionary run) obtained for each considered configuration. On the x-axis, we indicate the percentage of remaining working connections (after pruning) with respect to the total number of synapses, grouped every $10\%$, while on the y-axis we indicate how many networks (out of $30$, one per run) have that number of connections. For example, if we consider a point at $x = 40\%, y=0.5$, we mean that in $50\%$ of runs (i.e., $15$ out of $30$) networks after pruning have a number of working connections in the range $(30\%,40\%]$.

In both figures, we can observe a quite clear pattern: all the FFNN configurations use the majority of the connections available.
On the other hand, the distributions of working connections for the SBNN have two peaks: the first one occurs between $10\%$ and $20\%$ for the Mountain Car task and between $20\%$ and $40\%$ for the Lunar Lander one; the second peak is in common between the two tasks at around $90\%$. We visually analyzed all the networks and discovered that the SBNNs that compose the first peak have a structure like the one shown in \Cref{fig:ndrnn_pruning_case_b}, where all the hidden nodes are disconnected. Interestingly, the percentage of this kind of structures increases when $pt$ increases, as the first peak is higher for higher values of $pt$. This suggests that the later pruning occurs, the more probable it is that connections to the hidden nodes are pruned, thus leaving only input-output connections. Our intuition is that this form of simplification somehow correlates with the complexity needed to solve the task.

Concerning the total number of connections, we observe that in the Lunar Lander environment SBNNs maintain between $10$ and $30$ working connections for pruning rates higher than $20\%$. This range appears independent on the pruning time and the number of hidden nodes available.
On the contrary, FFNNs use the majority of the connections available, as for Mountain Car. Hence, in this case the number of working connections is strongly dependent on the number of hidden nodes and the pruning rate, resulting in a total number of working connections between $10$ and $64$.
The fact that the number of working connections varies is especially relevant when comparing the SBNN and the FFNN: for example, on the Lunar Lander task, the FFNN with $9$ hidden nodes and a $40\%$ pruning rate uses all the $64$ connections available, while the SBNN solves the task and obtains better performance using on average only $20$ connections, see \Cref{fig:ll_train}.

\begin{figure}[ht]
 \begin{tikzpicture}
 \node[scale=0.6] (S0) at (0,1) {\vt{\pic{MChn={}};}};
 \node[scale=0.6] (S0) at (5,1) {\vt{\pic{MCnhn={}};}};
 \end{tikzpicture}
 \caption{Structure of two selected SBNNs obtained on the Mountain Car task. On the left, an SBNN that uses $2$ out of $4$ hidden nodes and a $2$-layer structure. On the right, an SBNN that uses only two connections.}
 \label{fig:sbms}
\end{figure}
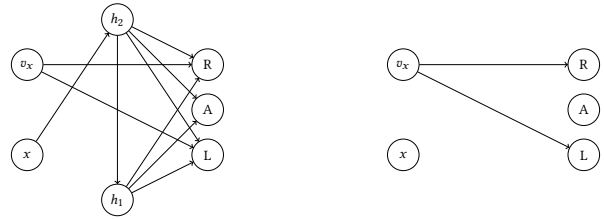

In \Cref{fig:sbms}, we show two SBNNs (after pruning) trained on the Mountain Car task. These networks are of two different kinds: one where all the hidden nodes have been removed, and that uses only two actions (accelerates left and accelerate right); and one that uses uses $2$ out of $4$ hidden nodes available, and all the three available actions.

\subsection{RQ3: Generalization}
In this section, we evaluate the generalization capabilities of the SBNN by testing the best agent (found after an evolution process on a given task) on another, unseen task. 

In particular, we test the agents trained on Lunar Lander on the Cart Pole and Mountain Car tasks. 
We restrict the validation to this case, as the Lunar Lander environment is the only one with enough input and output nodes to perform the other two tasks. In fact, the different tasks have different input and action spaces. 

To perform this analysis, we remap inputs and outputs from the validation task to the relative inputs in the Lunar Lander environment. For example, if in the Lunar Lander environment the first input is the x-position, we map the observation of the position from the validation task (i.e., the x-position of the cart or the car, respectively for Cart Pole and Mountain Car), to the first input. We set all the unused inputs to $0$. 
Regarding the output, we consider only the outputs present in the validation task. For example, considering the validation task as Cart Pole, the only actions available are \emph{move left} and \emph{move right}. Hence, we consider only the outputs that control the left and right engines in Lunar Lander.
 
\begin{figure}[ht]
 \centering
 \begin{tikzpicture}
 \begin{groupplot}[
 boxplot,
 boxplot/draw direction=y,
 group style={
 group size=2 by 3,
 horizontal sep=1mm,
 vertical sep=1.5mm,
 xticklabels at=edge bottom,
 yticklabels at=edge left
 },
 ymin=-10,
 ymax=610,
 ytick={0,250,500},
 width=0.5\columnwidth,
 height=0.5\columnwidth,
 legend cell align={center},
 xmajorticks=false
 ]
 \nextgroupplot[
 align=center,
 title={Cart Pole},
 ylabel = {Hidden nodes $5$ \\ Reward},
 legend columns=4,
 legend entries={PR 20, PR 60, PR 80},
 legend style={draw=none},
 legend to name=val
 ]
 \addlegendimage{area legend,color=cola1,fill}
 \addlegendimage{area legend,color=cola3,fill}
 \addlegendimage{area legend,color=cola4,fill}
 \addplot[black,fill=cola1] table[y=1_20_5] {data/cp_sbm.txt};
 \addplot[pattern={north east lines},pattern color=cola1] table[y=20_5] {data/cp_nn.txt};
 \pvalue{1}{2}{550}{25}{$\mathbf{<0.01}$}
 
 \addplot[black,fill=cola3] table[y=1_60_5] {data/cp_sbm.txt};
 \addplot[pattern={north east lines},pattern color=cola3] table[y=60_5] {data/cp_nn.txt};
 \pvalue{3}{4}{550}{25}{$\mathbf{0.04}$}
 
 \addplot[black,fill=cola4] table[y=1_80_5] {data/cp_sbm.txt};
 \addplot[pattern={north east lines},pattern color=cola4] table[y=80_5] {data/cp_nn.txt};
 \pvalue{5}{6}{550}{25}{$0.5$}
 
 \draw [dashed, red] (\pgfkeysvalueof{/pgfplots/xmin}, 475) -- (\pgfkeysvalueof{/pgfplots/xmax}, 475);
 \nextgroupplot[
 align=center,
 title={Mountain Car},
 legend columns=2,
 legend entries={SBNN, FFNN},
 legend style={draw=none},
 ymin=-210,
 ymax=-80,
 ytick={-210,-150,-110},
 yticklabels={$-210$,$-150$,$-110$},
 yticklabel pos=right,
 legend to name=val2
 ]
 \addlegendimage{area legend}
 \addlegendimage{area legend,pattern={north east lines},pattern color=black}
 \addplot[black,fill=cola1] table[y=1_20_5] {data/mc_sbm.txt};
 \addplot[pattern={north east lines},pattern color=cola1] table[y=20_5] {data/mc_nn.txt};
 \pvalue{1}{2}{-100}{5}{$0.7$}
 \addplot[black,fill=cola3] table[y=1_60_5] {data/mc_sbm.txt};
 \addplot[pattern={north east lines},pattern color=cola3] table[y=60_5] {data/mc_nn.txt};
 \pvalue{3}{4}{-100}{5}{$0.9$}
 \addplot[black,fill=cola4] table[y=1_80_5] {data/mc_sbm.txt};
 \addplot[pattern={north east lines},pattern color=cola4] table[y=80_5] {data/mc_nn.txt};
 \pvalue{5}{6}{-100}{5}{$\mathbf{<0.01}$}
 \draw [dashed, red] (\pgfkeysvalueof{/pgfplots/xmin}, -110) -- (\pgfkeysvalueof{/pgfplots/xmax}, -110);
 \nextgroupplot[
 align=center,
 ylabel = {Hidden nodes $7$ \\ Reward},
 ]
 \addplot[black,fill=cola1] table[y=1_20_7] {data/cp_sbm.txt};
 \addplot[pattern={north east lines},pattern color=cola1] table[y=20_7] {data/cp_nn.txt};
 \pvalue{1}{2}{550}{25}{$\mathbf{0.01}$}
 \addplot[black,fill=cola3] table[y=1_60_7] {data/cp_sbm.txt};
 \addplot[pattern={north east lines},pattern color=cola3] table[y=60_7] {data/cp_nn.txt};
 \pvalue{3}{4}{550}{25}{$\mathbf{0.02}$}
 \addplot[black,fill=cola4] table[y=1_80_7] {data/cp_sbm.txt};
 \addplot[pattern={north east lines},pattern color=cola4] table[y=80_7] {data/cp_nn.txt};
 \pvalue{5}{6}{550}{25}{$0.3$}
 
 \draw [dashed, red] (\pgfkeysvalueof{/pgfplots/xmin}, 475) -- (\pgfkeysvalueof{/pgfplots/xmax}, 475);
 \nextgroupplot[
 align=center,
 ymin=-210,
 ymax=-80,
 ytick={-210,-150,-110},
 yticklabels={$-210$,$-150$,$-110$},
 yticklabel pos=right,
 ]
 \addplot[black,fill=cola1] table[y=1_20_7] {data/mc_sbm.txt};
 \addplot[pattern={north east lines},pattern color=cola1] table[y=20_7] {data/mc_nn.txt};
 \pvalue{1}{2}{-100}{5}{$0.13$}
 \addplot[black,fill=cola3] table[y=1_60_7] {data/mc_sbm.txt};
 \addplot[pattern={north east lines},pattern color=cola3] table[y=60_7] {data/mc_nn.txt};
 \pvalue{3}{4}{-100}{5}{$0.7$}
 \addplot[black,fill=cola4] table[y=1_80_7] {data/mc_sbm.txt};
 \addplot[pattern={north east lines},pattern color=cola4] table[y=80_7] {data/mc_nn.txt};
 \pvalue{5}{6}{-100}{5}{$0.16$}
 \draw [dashed, red] (\pgfkeysvalueof{/pgfplots/xmin}, -110) -- (\pgfkeysvalueof{/pgfplots/xmax}, -110);
 \nextgroupplot[
 align=center,
 ylabel = {Hidden nodes $9$ \\ Reward},
 ]
 \addplot[black,fill=cola1] table[y=1_20_9] {data/cp_sbm.txt};
 \addplot[pattern={north east lines},pattern color=cola1] table[y=20_9] {data/cp_nn.txt};
 \pvalue{1}{2}{550}{25}{$\mathbf{<0.01}$}
 \addplot[black,fill=cola3] table[y=1_60_9] {data/cp_sbm.txt};
 \addplot[pattern={north east lines},pattern color=cola3] table[y=60_9] {data/cp_nn.txt};
 \pvalue{3}{4}{550}{25}{$\mathbf{<0.01}$}
 \addplot[black,fill=cola4] table[y=1_80_9] {data/cp_sbm.txt};
 \addplot[pattern={north east lines},pattern color=cola4] table[y=80_9] {data/cp_nn.txt};
 \pvalue{5}{6}{550}{25}{$\mathbf{0.04}$}
 \draw [dashed, red] (\pgfkeysvalueof{/pgfplots/xmin}, 475) -- (\pgfkeysvalueof{/pgfplots/xmax}, 475);
 \nextgroupplot[
 align=center,
 ymin=-210,
 ymax=-80,
 ytick={-210,-150,-110},
 yticklabels={$-210$,$-150$,$-110$},
 yticklabel pos=right,
 ]
 \addplot[black,fill=cola1] table[y=1_20_9] {data/mc_sbm.txt};
 \addplot[pattern={north east lines},pattern color=cola1] table[y=20_9] {data/mc_nn.txt};
 \pvalue{1}{2}{-100}{5}{$0.25$}
 \addplot[black,fill=cola3] table[y=1_60_9] {data/mc_sbm.txt};
 \addplot[pattern={north east lines},pattern color=cola3] table[y=60_9] {data/mc_nn.txt};
 \pvalue{3}{4}{-100}{5}{$0.62$}
 \addplot[black,fill=cola4] table[y=1_80_9] {data/mc_sbm.txt};
 \addplot[pattern={north east lines},pattern color=cola4] table[y=80_9] {data/mc_nn.txt};
 \pvalue{5}{6}{-100}{5}{$0.14$}
 \draw [dashed, red] (\pgfkeysvalueof{/pgfplots/xmin}, -110.0) -- (\pgfkeysvalueof{/pgfplots/xmax}, -110.0);
 \end{groupplot}
 \end{tikzpicture}\\
 \pgfplotslegendfromname{val}\\
 \pgfplotslegendfromname{val2}
 \caption{Comparison between the SBNN with $pt=1$ and the FFNN trained on the Lunar Lander task. The y-axis represents the average reward in the two validation tasks: Cart Pole (left) and Mountain Car (right). The dashed red line indicates the solving threshold for the two tasks. The results show that SBNN trained on the Lunar Lander task tends to achieve better or equal performance with respect to the FFNN. For each pairwise comparison, we indicate the p-value of the Wilcoxon rank-sum test.}
 \label{fig:val}
\end{figure}
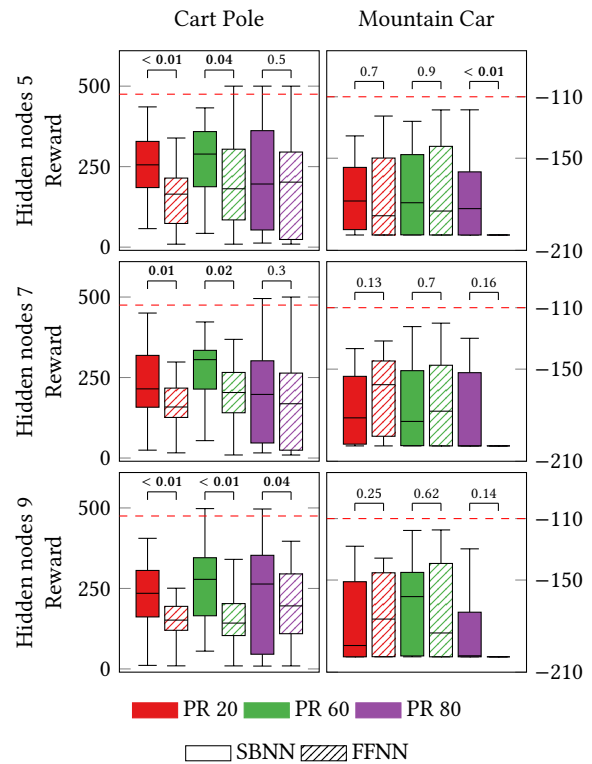

\Cref{fig:val} shows the performances in the validation tasks for a subset of the configurations (statistical significance assessed with Wilcoxon rank-sum test, $\alpha = 0.05$). On the Cart Pole task, the results of the SBNN are similar to or better than the ones of the FFNN. In the Mountain Car environment, the SBNN performs slightly worse than the FFNN for $pr=20\%$, although the differences are not always statistically significant; for $pr=80\%$, the SBNN shows better performances, with higher performance in the first quartile (indicating, once again, that the SBNN models effectively learns to use the network structure).

Finally, with these validation experiments, we can observe how the task affects the network structure. \Cref{fig:brain_cmp} shows the network structure of the same agent, which can solve both the Lunar Lander and the Cart Pole task after the pruning process. We can observe that the networks differ in the number of connections used and in the neurons that those synapses connect.

 \begin{figure}[ht]
 \begin{tikzpicture}
 \node[scale=0.5] (S0) at (0,1) {\vt{\pic{LLT={}};}};
 \node[scale=0.5] (S0) at (5,1) {\vt{\pic{LLCP={}};}};
 \end{tikzpicture}
 \caption{Structure of the same SBNN in the Lunar Lander (left) and Cart Pole (right) environments. We can observe how the network structure changes based on the task.}
 \label{fig:brain_cmp}
\end{figure}
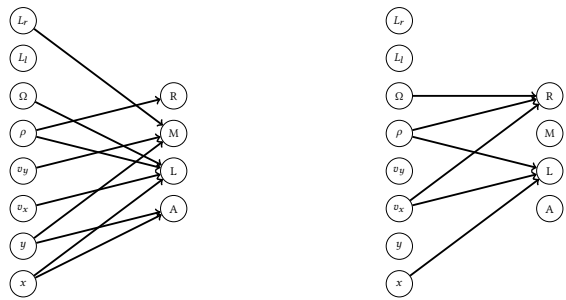
 
\section{Conclusions}
\label{sec:conclusions}
In this work, we took inspiration from the synaptogenesis process occurring in natural brains to propose a new learning model that combines both plasticity and pruning.
At the beginning of the first episode of the task, all the neurons are connected with each other and initialized to $0$ (i.e., the connections exist, but they are initially deactivated).
Then, within each episode, a plasticity model based on Hebbian learning grows those synapses, activating the corresponding connections. Eventually, during the life of the agent (i.e. at a predetermined episode), those connections are pruned through a global magnitude pruning algorithm and Hebbian learning is stopped.
We called this model Self-building Neural Network (SBNN), as it changes its structure based on the experience of the agent during the episodes of the task.

We tested our model on three classical control tasks from the OpenAI, namely Cart Pole, Mountain Car, and Lunar Lander.
In our experiments, we varied the three main parameters of the model, affecting respectively \emph{when} to prune, \emph{how much} to prune, and the number of hidden nodes. 
We showed that, in general, the SBNN reaches better performance than the FFNN, and that it can adapt better to unseen tasks. Furthermore, we assessed the importance of the model's parameters, in particular regarding when and how much to prune. 
Finally, we highlighted how the same agent reorganizes its brain differently, based on the task, and how it can remove unnecessary complexity from the brain, given enough time.

In the future, we plan to develop a system for automatically deciding how much and when to prune based on the information flow in the network. 
In addition, we aim to tackle more complex tasks, such as the control of soft robots, where we can test the proposed SBNN on larger input and action spaces.
We also plan to address the two limitations indicated in \Cref{sec:methodology}. For the number of parameters to optimize, we will test the use of a Hebbian rule for each neuron, rather than one for each connection, also modifying the update rule to avoid that each connection receives the same update. 
For the activation order, we plan to use a distance-based approach to solve the cycle, using as the distance the weights.

\bibliographystyle{ACM-Reference-Format}
\bibliography{main}

\end{document}

%% file: drawing.tex
\usetikzlibrary{positioning}
\newcommand{\vt}[1]{\adjustbox{}{\tikz{#1}}}

\newcommand{\sensors}[9]{ 
    \ifthenelse{#3 = 100}{
        \filldraw[fill=red!#3, draw=black!#3] (\l/5+#1*\l -\l/15 , \l/2+#2*\l - \l/15) rectangle (\l/5+#1*\l +\l/15 , \l/2+#2*\l + \l/15);
    }{
    \pgfmathsetmacro{\cnt}{#3+ 50}
        \filldraw[fill=black!0, draw=black!\cnt] (\l/5+#1*\l -\l/15 , \l/2+#2*\l - \l/15) rectangle (\l/5+#1*\l +\l/15 , \l/2+#2*\l + \l/15);
    }
    \ifthenelse{#4 = 100}{
        \filldraw[fill=red!#4, draw=black!#4]  (2*\l/5+#1*\l-\l/15, \l/2+#2*\l-\l/15) rectangle (2*\l/5+#1*\l+\l/15, \l/2+#2*\l+\l/15);
    }{
    \pgfmathsetmacro{\cnt}{#4+ 50}
        \filldraw[fill=black!0, draw = black!\cnt] (2*\l/5+#1*\l-\l/15, \l/2+#2*\l-\l/15) rectangle (2*\l/5+#1*\l+\l/15, \l/2+#2*\l+\l/15);
    }
    \ifthenelse{#5 = 100}{
        \filldraw[fill=red!#5, draw=black!#5] (3*\l/5+#1*\l-\l/15, \l/2+#2*\l) rectangle (3*\l/5+#1*\l+\l/15, \l/2+#2*\l+2*\l/15);
    }{
    \pgfmathsetmacro{\cnt}{#5+ 50}
        \filldraw[fill=black!0, draw=black!\cnt] (3*\l/5+#1*\l-\l/15, \l/2+#2*\l) rectangle (3*\l/5+#1*\l+\l/15, \l/2+#2*\l+2*\l/15);
    }
    \ifthenelse{#6 = 100}{
        \filldraw[fill=red!#6, draw=black!#6] (3*\l/5+#1*\l-\l/15, \l/2+#2*\l) rectangle (3*\l/5+#1*\l+\l/15, \l/2+#2*\l-2*\l/15);
    }{
    \pgfmathsetmacro{\cnt}{#6+ 50}
       \filldraw[fill=black!0, draw=black!\cnt] (3*\l/5+#1*\l-\l/15, \l/2+#2*\l) rectangle (3*\l/5+#1*\l+\l/15, \l/2+#2*\l-2*\l/15);
    }
    \ifthenelse{#7 = 100}{
        \filldraw[fill=red!#7, draw=black!#7] (4*\l/5+#1*\l-\l/15, \l/2+#2*\l+\l/15) rectangle (4*\l/5+#1*\l+\l/15, \l/2+#2*\l+\l/5); 
    }{
    \pgfmathsetmacro{\cnt}{#7+ 50}
        \filldraw[fill=black!0, draw=black!\cnt] (4*\l/5+#1*\l-\l/15, \l/2+#2*\l+\l/15) rectangle (4*\l/5+#1*\l+\l/15, \l/2+#2*\l+\l/5); 
    }
    \ifthenelse{#8 = 100}{
        \filldraw[fill=red!#8, draw=black!#8] (4*\l/5+#1*\l-\l/15, \l/2+#2*\l-\l/15) rectangle (4*\l/5+#1*\l+\l/15, \l/2+#2*\l+\l/15);
    }{
    \pgfmathsetmacro{\cnt}{#8+ 50}
        \filldraw[fill=black!0, draw=black!\cnt] (4*\l/5+#1*\l-\l/15, \l/2+#2*\l-\l/15) rectangle (4*\l/5+#1*\l+\l/15, \l/2+#2*\l+\l/15);
    }
    \ifthenelse{#9 = 100}{
        \filldraw[fill=red!#9, draw=black!#9] (4*\l/5+#1*\l-\l/15, \l/2+#2*\l-\l/15) rectangle (4*\l/5+#1*\l+\l/15, \l/2+#2*\l-\l/5);
    }{
        \pgfmathsetmacro{\cnt}{#9 + 50}
        \filldraw[fill=black!0, draw=black!\cnt] (4*\l/5+#1*\l-\l/15, \l/2+#2*\l-\l/15) rectangle (4*\l/5+#1*\l+\l/15, \l/2+#2*\l-\l/5);
    }
}

\newcommand{\ndrcasec}[0]{
    \node[circle, draw] (I) at (0,1){I}; 
    
    \node[circle, draw] (A) at (1,0){A};
    \node[circle, draw] (B) at (1,1){B};
    \node[circle, draw] (C) at (1,2){C};
    \node[circle, draw] (O) at (2,1){O};
    
    \node[circle] (D) at (1,3){};
    
    \draw[->] (I) to[right] node[left]{}(B) ;
    \draw[->] (I) to[right] node[left]{}(A) ;
    \draw[->, red] (I) to[right] node[left]{}(C) ;
    
    \draw[->, red] (A) to[right] node[left]{}(O) ;
    \draw[->, red] (B) to[right] node[left]{}(O) ;
    \draw[->] (C) to[right] node[left]{}(O) ;
    
    \draw[<->, red] (A) to[right] node[left]{}(B) ;
    \draw[->] (B) to[right] node[left]{}(C) ;
    \draw[->] (A) to[bend right] node[left]{}(C) ;
    
    \draw[-] (I) to[bend left=60] node[left]{}(D.west);
    \draw[-] (D.east) to node[left]{}(D.west);
    \draw[->] (D.east) to[bend left=60] node[left]{}(O);
}
\tikzset{pics/ndrcasecpc/.style n args={2}{code={
    \ndrcasec
}}}

\newcommand{\ndrcaseb}[0]{
    \node[circle, draw] (I) at (0,1){I}; 
    
    \node[circle, draw] (A) at (1,0){A};
    \node[circle, draw] (B) at (1,1){B};
    \node[circle, draw] (C) at (1,2){C};
    \node[circle, draw] (O) at (2,1){O};
    
    \node[circle] (D) at (1,3){};
    
    \draw[->, red] (I) to[right] node[left]{}(B) ;
    \draw[->, red] (I) to[right] node[left]{}(A) ;
    \draw[->, red] (I) to[right] node[left]{}(C) ;
    
    \draw[->, red] (A) to[right] node[left]{}(O) ;
    \draw[->, red] (B) to[right] node[left]{}(O) ;
    \draw[->, red] (C) to[right] node[left]{}(O) ;
    
    \draw[<->, red] (A) to[right] node[left]{}(B) ;
    \draw[<->, red] (B) to[right] node[left]{}(C) ;
    \draw[<->, red] (A) to[bend right] node[left]{}(C) ;
    
    \draw[-] (I) to[bend left=60] node[left]{}(D.west);
    \draw[-] (D.east) to node[left]{}(D.west);
    \draw[->] (D.east) to[bend left=60] node[left]{}(O);
}
\tikzset{pics/ndrcasebpc/.style n args={2}{code={
    \ndrcaseb
}}}

\newcommand{\ndrcasea}[0]{
    \node[circle, draw] (I) at (0,1){I}; 
    
    \node[circle, draw] (A) at (1,0){A};
    \node[circle, draw] (B) at (1,1){B};
    \node[circle, draw] (C) at (1,2){C};
    \node[circle, draw] (O) at (2,1){O};
    
    \node[circle] (D) at (1,3){};
    
    \draw[->] (I) to[right] node[left]{}(B) ;
    \draw[->] (I) to[right] node[left]{}(A) ;
    \draw[->] (I) to[right] node[left]{}(C) ;
    
    \draw[->] (A) to[right] node[left]{}(O) ;
    \draw[->] (B) to[right] node[left]{}(O) ;
    \draw[->] (C) to[right] node[left]{}(O) ;
    
    \draw[<->, red] (A) to[right] node[left]{}(B) ;
    \draw[<->, red] (B) to[right] node[left]{}(C) ;
    \draw[<->, red] (A) to[bend right] node[left]{}(C) ;
    
    \draw[-] (I) to[bend left=60] node[left]{}(D.west);
    \draw[-] (D.east) to node[left]{}(D.west);
    \draw[->] (D.east) to[bend left=60] node[left]{}(O);
}

\tikzset{pics/ndrcaseapc/.style n args={2}{code={
    \ndrcasea
}}}

\tikzset{pics/ndrStart/.style n args={2}{code={
        \node[circle, draw] (I) at (#1 +0,#2 +1){I}; 
        \node[circle, draw] (A) at (#1 +1,#2){A};
        \node[circle, draw] (B) at (#1 +1,#2 +1){B};
        \node[circle, draw] (C) at (#1 +1,#2 +2){C};
        \node[circle, draw] (O) at (#1 +2,#2 +1){O};
        
        \node[circle] (D) at (#1 + 1,#2 +3){};
        
        \draw[->, red] (I) to[right] node[left]{}(B) ;
        \draw[->, red] (I) to[right] node[left]{}(A) ;
        \draw[->, red] (I) to[right] node[left]{}(C) ;
        
        \draw[->, red] (A) to[right] node[left]{}(O) ;
        \draw[->, red] (B) to[right] node[left]{}(O) ;
        \draw[->, red] (C) to[right] node[left]{}(O) ;
    
        \draw[<->, red] (A) to[right] node[left]{}(B) ;
        \draw[<->, red] (B) to[right] node[left]{}(C) ;%
        \draw[<->, red] (A) to[bend right] node[left]{}(C) ;
        
        \draw[-, red] (I) to[bend left=60] node[left]{}(D.west);
        \draw[-, red] (D.east) to node[left]{}(D.west);
        \draw[->, red] (D.east) to[bend left=60] node[left]{}(O);
}}}

\tikzset{pics/ndr1step/.style n args={2}{code={
        \node[circle, draw] (I) at (#1 +0,#2 +1){I}; 
        \node[circle, draw] (A) at (#1 +1,#2){A};
        \node[circle, draw] (B) at (#1 +1,#2 +1){B};
        \node[circle, draw] (C) at (#1 +1,#2 +2){C};
        \node[circle, draw] (O) at (#1 +2,#2 +1){O};
        
        \node[circle] (D) at (#1 + 1,#2 +3){};
        
        \draw[->, line width=0.5mm] (I) to[right] node[left]{}(B) ;
        \draw[->, line width=0.12mm] (I) to[right] node[left]{}(A) ;
        \draw[->, line width=0.15mm] (I) to[right] node[left]{}(C) ;
        
        \draw[->, line width=0.6mm] (A) to[right] node[left]{}(O) ;
        \draw[->, line width=0.02mm] (B) to[right] node[left]{}(O) ;
        \draw[->, line width=0.6mm] (C) to[right] node[left]{}(O) ;
    
        \draw[<->, line width=0.35mm] (A) to[right] node[left]{}(B) ;
        \draw[<->, line width=0.43mm] (B) to[right] node[left]{}(C) ;%
        \draw[<->, line width=0.23mm] (A) to[bend right] node[left]{}(C) ;
        
        \draw[-, line width=0.03mm] (I) to[bend left=60] node[left]{}(D.west);
        \draw[-, line width=0.03mm] (D.east) to node[left]{}(D.west);
        \draw[->, line width=0.03mm] (D.east) to[bend left=60] node[left]{}(O);
}}}
        
\tikzset{pics/ndrpp/.style n args={2}{code={
        \node[circle, draw] (I) at (#1 +0,#2 +1){I}; 
        \node[circle, draw] (A) at (#1 +1,#2){A};
        \node[circle, draw] (B) at (#1 +1,#2 +1){B};
        \node[circle, draw] (C) at (#1 +1,#2 +2){C};
        \node[circle, draw] (O) at (#1 +2,#2 +1){O};
        
        \node[circle] (D) at (#1 + 1,#2 +3){};
        
        \draw[->, line width=0.5mm] (I) to[right] node[left]{}(B) ;
        \draw[->, red] (I) to[right] node[left]{}(A) ;
        \draw[->, red] (I) to[right] node[left]{}(C) ;
        
        \draw[->, line width=0.6mm] (A) to[right] node[left]{}(O) ;
        \draw[->, red] (B) to[right] node[left]{}(O) ;
        \draw[->, line width=0.6mm] (C) to[right] node[left]{}(O) ;
    
        \draw[<->, line width=0.35mm] (A) to[right] node[left]{}(B) ;
        \draw[<->, line width=0.43mm] (B) to[right] node[left]{}(C) ;%
        \draw[<->, line width=0.23mm] (A) to[bend right] node[left]{}(C) ;
        
        \draw[-, red] (I) to[bend left=60] node[left]{}(D.west);
        \draw[-, red] (D.east) to node[left]{}(D.west);
        \draw[->, red] (D.east) to[bend left=60] node[left]{}(O);
}}}

\tikzset{pics/hier/.style n args={1}{code={
            \def\l{#1} 
            \foreach \x in {0,1,2,3,4,5}
                \draw (0+\l,0+\x) rectangle (\l+\l,\l+\x) node[pos=.5] {$i_{\x}$};
            \draw [decorate, decoration = {calligraphic brace, mirror,amplitude=10pt}] (2.5,4.1) --  (2.5,5.9);
            \draw [decorate, decoration = {calligraphic brace, mirror,amplitude=10pt}] (2.5,2.1) --  (2.5,3.9);
            \draw [decorate, decoration = {calligraphic brace, mirror,amplitude=10pt}] (2.5,0.1) --  (2.5,1.9);
            
            \node[rectangle, draw] (a) at (3.5, 1){$SBM_{0}$};
            \node[rectangle, draw] (b) at (3.5, 3){$SBM_{1}$};
            \node[rectangle, draw] (c) at (3.5, 5){$SBM_{2}$};

            \node[rectangle, draw,align=center] (head) at (6.5, 3) {$Interpretable$ \\ $Model$};
            \node[rectangle, draw] (leaf2) at (9, 3) {$output$};
            
            \draw[->, shorten <=2pt,shorten >=2pt,>=stealth] (a.east) to[right] node[left]{}(head.west) ;
            \draw[->, shorten <=2pt,shorten >=2pt,>=stealth] (b.east) to[right] node[left]{}(head.west) ;
            \draw[->, shorten <=2pt,shorten >=2pt,>=stealth] (c.east) to[right] node[left]{}(head.west) ;
            
            \draw[->, shorten <=2pt,shorten >=2pt,>=stealth] (head) to[right] node[left]{}(leaf2.west) ;
}}}

\tikzset{pics/treeRep/.style n args={1}{code={
            \def\l{1} 
            \foreach \x in {0,1,2,3,4,5}
                \draw (0+\x*\l,0) rectangle (\l+\x*\l,\l) node[pos=0.5]{$\x$};
            \draw [decorate, decoration = {calligraphic brace, mirror,amplitude=10pt}] (0.1,-0.5) --  (5.9,-0.5) node[pos=0.5, below=10pt]{Genome};
            \draw[->] (6.6, 0.5) -- node[above=10pt,align=center, pos=0.5]{translation \\  process} (10.5, 0.5);
            
            \node[rectangle, draw] (head) at (14,2.5) {Condition};
            \node[rectangle, draw] (condition1) at (12.5,0.5) {Condition};
            \node[circle, draw] (leaf1) at (15.5,0.5) {RL};
            \node[circle, draw] (leaf2) at (11.5,-1.5) {RL};
            \node[circle, draw] (leaf3) at (13.5,-1.5) {RL};
            
            \draw[->, shorten <=2pt,shorten >=2pt,>=stealth] (head.south) to[below] node[above]{}(condition1.north) ;
            \draw[->, shorten <=2pt,shorten >=2pt,>=stealth] (head.south) to[below] node[above]{}(leaf1.north) ;
            \draw[->, shorten <=2pt,shorten >=2pt,>=stealth] (condition1.south) to[below] node[above]{}(leaf2.north) ;
            \draw[->, shorten <=2pt,shorten >=2pt,>=stealth] (condition1.south) to[below] node[above]{}(leaf3.north) ;
}}}

\tikzset{pics/gmm/.style n args={0}{code={
        \draw[yscale=0.5,yshift=-3, domain=0:10,blue] plot function{exp(-(x-5)*(x-5)/2/2/2.0)/sqrt(pi)/2};
}}}

\tikzset{pics/vsrRep/.style n args={0}{code={
        \def\l{1}
        \foreach \x in {0,1,2,3,4,5}
            \draw (0+\x*\l,0) rectangle (\l+\x*\l,\l) node[pos=0.5]{$0.\x$};
        \draw [decorate, decoration = {calligraphic brace, mirror,amplitude=10pt}] (0.1,-0.5) --  (5.9,-0.5) node[pos=0.5, below=10pt]{Genome};
        
        \draw[-] (6.6, 0.5) -- node[above=10pt,align=center, pos=0.5]{translation \\  process} (9.5, 0.5);
        \draw[-] (9.5,1.5) -- (9.5,-0.5);
        \draw[->] (9.5,-0.5) -- (10.5,-0.5);
        \draw[->] (9.5,1.5) -- (10.5,1.5);

        \node[draw,xscale=0.1] (BG) at (11.5,1.5) {\vt{\pic{gmm={}};}}; 
        \node[below= 0.1pt of BG,align=center]  {body \\ generator};
        
        \node[draw,xscale=0.1] (BC) at (11.5,-0.5) {\vt{\pic{gmm={}};}}; 
        \node[below= 0.1pt of BC,align=center]  {controller \\ generator};
        
        \node[] (r) at (14.5, 0.5) {\vsr[2mm]{7}{6}{0011110-0011010-1111111-0111110-0100110-0110010}};
        
        \draw[->, shorten <=2pt,shorten >=2pt,>=stealth] (BG.east) to node[]{}(r.west) ;
        \draw[->, shorten <=2pt,shorten >=2pt,>=stealth] (BC.east) to node[]{}(r.west) ;
}}}

\tikzset{pics/direct/.style n args={0}{code={
 \draw [->](-0,0) -- node[above, xshift=5mm, yshift=1mm, align=center] {Model\\complexity} ++ (5, 0);
 \draw (2, -0.25) -- node[below, yshift=-2mm] {Single} ++ (0, 0.5);
 \draw (4, -0.25) -- node[below, yshift=-2mm] {Multiple} ++ (0, 0.5);

 \draw [->](1, 1) -- node[left, yshift=-5mm, align=center] {Task\\complexity} ++ (0,-4);
 \draw (0.75, -1) -- node[right, xshift=2mm] {OpenAI} ++ (0.5, 0);
 \draw (0.75, -2.25) -- node[right, xshift=2mm] {VSRs} ++ (0.5, 0);
}}}

\newcommand{\ndrCycle}[1]{
    \node[circle, draw] (I) at (0+#1,1){I};

    \node[circle, draw] (C) at (1+#1,2){C};
    \node[circle, draw] (B) at (1+#1,1){B};
    \node[circle, draw] (A) at (1+#1,0){A};

    \node[circle, draw] (O) at (2+#1,1){O};

    \draw[->] (I) to[right] node[left]{}(A) ;
    \draw[->] (I) to[right] node[left]{}(B) ;
    \draw[->] (I) to[right] node[left]{}(C) ;

    \draw[<->] (A) to[right] node[left]{}(B) ;
    \draw[<->] (C) to[right] node[left]{}(B) ;

    \draw[->] (A) to[right] node[left]{}(O) ;
    \draw[->] (B) to[right] node[left]{}(O) ;
    \draw[->] (C) to[right] node[left]{}(O) ;
    
}

\newcommand{\ndrCycleHL}[1]{
    \ndrCycle{#1}
    \draw[red] (1+#1,1.5) ellipse (0.5cm and 1cm);

}
\newcommand{\ndrCycleFK}[1]{
    \node[circle, draw] (I) at (0+#1,1){I};

    \node[circle, draw, red] (F1) at (1+#1,2){F1};
    \node[circle, draw] (A) at (1+#1,0){A};

    \node[circle, draw] (O) at (2+#1,1){O};

    \draw[->] (I) to[right] node[left]{}(F1) ;
    \draw[->] (I) to[right] node[left]{}(A) ;

    \draw[<->] (F1) to[right] node[left]{}(A) ;
    \draw[<->] (A) to[right] node[left]{}(F1) ;

    \draw[->] (F1) to[right] node[left]{}(O) ;
    \draw[->] (A) to[right] node[left]{}(O) ;
}
\newcommand{\ndrCycleFHL}[1]{
    \ndrCycleFK{+#1}
    \draw[blue] (1+#1,1) ellipse (0.5cm and 1.5cm);

}

\newcommand{\ndrCycleFKp}[1]{
    \node[circle, draw] (I) at (0+#1,1){I};

    \node[circle, draw, blue] (F2) at (1+#1,1){F2};

    \node[circle, draw] (O) at (2+#1,1){O};

    \draw[->] (I) to[right] node[left]{}(F2) ;

    \draw[->] (F2) to[right] node[left]{}(O) ;
}

\tikzset{pics/ndrCyclepc/.style n args={1}{code={
    \ndrCycle{#1}
}}}

\tikzset{pics/ndrCycleHLpc/.style n args={1}{code={
    \ndrCycleHL{#1}
}}}
\tikzset{pics/ndrCycleFKpc/.style n args={1}{code={
    \ndrCycleFK{#1}
}}}
\tikzset{pics/ndrCycleFHLpc/.style n args={1}{code={
    \ndrCycleFHL{#1}
}}}
\tikzset{pics/ndrCycleFKppc/.style n args={1}{code={
    \ndrCycleFKp{#1}
}}}
\tikzset{pics/ndrCyclepc/.style n args={1}{code={
    \ndrCycle{#1}
}}}

\tikzset{pics/LLT/.style n args={0}{code={
        \node[circle, draw, minimum size=0.7cm] (i0) at (0,0){$x$}; 
        \node[circle, draw, minimum size=0.7cm] (i1) at (0,1){$y$};
        \node[circle, draw, minimum size=0.7cm] (i2) at (0,2){$v_x$};
        \node[circle, draw, minimum size=0.7cm] (i3) at (0,3){$v_y$};
        \node[circle, draw, minimum size=0.7cm] (i4) at (0,4){$\rho$};
        \node[circle, draw, minimum size=0.7cm] (i5) at (0,5){$\Omega$};
        \node[circle, draw, minimum size=0.7cm] (i6) at (0,6){$L_l$};
        \node[circle, draw, minimum size=0.7cm] (i7) at (0,7){$L_r$};
        
        \node[circle, draw, minimum size=0.7cm] (o0) at (4,2){A}; 
        \node[circle, draw, minimum size=0.7cm] (o1) at (4,3){L};
        \node[circle, draw, minimum size=0.7cm] (o2) at (4,4){M};
        \node[circle, draw, minimum size=0.7cm] (o3) at (4,5){R};

        \draw[->, line width=0.5mm] (i0) to[right] node[left]{}(o0) ;
        \draw[->, line width=0.5mm] (i1) to[right] node[left]{}(o0) ;

        \draw[->, line width=0.5mm] (i0) to[right] node[left]{}(o1) ;
        \draw[->, line width=0.5mm] (i2) to[right] node[left]{}(o1) ;
        \draw[->, line width=0.5mm] (i4) to[right] node[left]{}(o1) ;
        \draw[->, line width=0.5mm] (i5) to[right] node[left]{}(o1) ;

        \draw[->, line width=0.5mm] (i1) to[right] node[left]{}(o2) ;
        \draw[->, line width=0.5mm] (i3) to[right] node[left]{}(o2) ;
        \draw[->, line width=0.5mm] (i7) to[right] node[left]{}(o2) ;

        \draw[->, line width=0.5mm] (i4) to[right] node[left]{}(o3) ;
}}}

\tikzset{pics/LLCP/.style n args={0}{code={
        \node[circle, draw, minimum size=0.7cm] (i0) at (0,0){$x$}; 
        \node[circle, draw, minimum size=0.7cm] (i1) at (0,1){$y$};
        \node[circle, draw, minimum size=0.7cm] (i2) at (0,2){$v_x$};
        \node[circle, draw, minimum size=0.7cm] (i3) at (0,3){$v_y$};
        \node[circle, draw, minimum size=0.7cm] (i4) at (0,4){$\rho$};
        \node[circle, draw, minimum size=0.7cm] (i5) at (0,5){$\Omega$};
        \node[circle, draw, minimum size=0.7cm] (i6) at (0,6){$L_l$};
        \node[circle, draw, minimum size=0.7cm] (i7) at (0,7){$L_r$};
        
        \node[circle, draw, minimum size=0.7cm] (o0) at (4,2){A}; 
        \node[circle, draw, minimum size=0.7cm] (o1) at (4,3){L};
        \node[circle, draw, minimum size=0.7cm] (o2) at (4,4){M};
        \node[circle, draw, minimum size=0.7cm] (o3) at (4,5){R};

        \draw[->, line width=0.5mm] (i0) to[right] node[left]{}(o1) ;
        \draw[->, line width=0.5mm] (i2) to[right] node[left]{}(o1) ;
        \draw[->, line width=0.5mm] (i4) to[right] node[left]{}(o1) ;

        \draw[->, line width=0.5mm] (i5) to[right] node[left]{}(o3) ;
        \draw[->, line width=0.5mm] (i2) to[right] node[left]{}(o3) ;
        \draw[->, line width=0.5mm] (i4) to[right] node[left]{}(o3) ;

}}}

\tikzset{pics/MCnhn/.style n args={0}{code={
        \node[circle, draw, minimum size=0.7cm] (i0) at (0,0){$x$}; 
        \node[circle, draw, minimum size=0.7cm] (i1) at (0,2){$v_x$};

        \node[circle, draw, minimum size=0.7cm] (o0) at (4,0){L};
        \node[circle, draw, minimum size=0.7cm](o1) at (4,1){A}; 
        \node[circle, draw, minimum size=0.7cm] (o2) at (4,2){R};

        \draw[->] (i1) to[right] node[left]{}(o0) ;
        \draw[->] (i1) to[right] node[left]{}(o2) ;

}}}

\tikzset{pics/MChn/.style n args={0}{code={
        \node[circle, draw, minimum size=0.7cm] (i0) at (0,0){$x$}; 
        \node[circle, draw, minimum size=0.7cm] (i1) at (0,2){$v_x$};
       
        \node[circle, draw, minimum size=0.7cm] (o0) at (4,0){L};
        \node[circle, draw, minimum size=0.7cm] (o1) at (4,1){A}; 
        \node[circle, draw, minimum size=0.7cm] (o2) at (4,2){R};
        
        \draw[->] (i1) to[right] node[left]{}(o0) ;
        \draw[->] (i1) to[right] node[left]{}(o2) ;
        
        \node[circle, draw, minimum size=0.7cm](h0) at (2,-1){$h_1$};
        \node[circle, draw, minimum size=0.7cm](h1) at (2,3){$h_2$};

        \draw[->] (i0) to[right] node[left]{}(h1) ;
        
        \draw[->] (h1) to[right] node[left]{}(h0) ;
        \draw[->] (h1) to[right] node[left]{}(o0) ;
        \draw[->] (h1) to[right] node[left]{}(o1) ;
        \draw[->] (h1) to[right] node[left]{}(o2) ;

        \draw[->] (h0) to[right] node[left]{}(o0) ;
        \draw[->] (h0) to[right] node[left]{}(o1) ;
        \draw[->] (h0) to[right] node[left]{}(o2) ;

}}}

%% file: plot-macros.tex
\newcommand{\linesimple}[4]{
    \addplot [each nth point=1,mark=*,mark size=1.5pt, #4,thick] table [x={#2},y={#3}] {#1};
}
\newcommand{\linesimpledash}[4]{
    \addplot [dashed, mark=*, mark size=1.5pt, each nth point=1, #4,thick] table [x={#2},y={#3}] {#1};
}
\newcommand{\linevb}[5]{
    \addplot [each nth point=1,mark=*, #5,thick, error bars/.cd, y dir=both, y explicit, error mark options={ rotate=90, #5, mark size=3pt}] table [x={#2},y={#3}, y error={#4}] {#1};
}
\newcommand{\linesdashvb}[5]{
    \addplot [dashed, mark=*, each nth point=1, #5, thick,  error bars/.cd, y dir=both, y explicit, error mark options={ rotate=90, #5, mark size=3pt}] table [x={#2},y={#3}, y error={#4}] {#1};
}

\newcommand{\pvalue}[5]{
    \draw (#1,#3) -- (#1,#3-#4);
    \draw (#2,#3) -- (#2,#3-#4);
    \draw (#1,#3) -- node [above,scale=.65] {#5} (#2,#3);
}

\def\addlegendimage{\csname pgfplots@addlegendimage\endcsname}

\definecolor{cola1}{RGB}{228,26,28}
\definecolor{cola2}{RGB}{55,126,184}
\definecolor{cola3}{RGB}{77,175,74}
\definecolor{cola4}{RGB}{152,78,163}
\definecolor{cola5}{RGB}{255,127,0}
\definecolor{cola6}{RGB}{255,255,51}
\definecolor{cola7}{RGB}{166,86,40}

\definecolor{colb1}{RGB}{102,194,165}
\definecolor{colb2}{RGB}{252,141,98}
\definecolor{colb3}{RGB}{141,160,203}
\definecolor{colb4}{RGB}{231,138,195}

\definecolor{colb1}{RGB}{102,194,165}
\definecolor{colb2}{RGB}{252,141,98}
\definecolor{colb3}{RGB}{141,160,203}
\definecolor{colb4}{RGB}{231,138,195}

\pgfplotsset{/pgfplots/colormap={evolvability}{rgb255=(241,163,64) rgb255=(224,224,224) rgb255=(153,142,195)}}
\pgfplotsset{/pgfplots/colormap={fitness}{rgb255=(229,245,249) rgb255=(153,216,201) rgb255=(44,162,95)}}
\pgfkeys{/pgf/number format/1000 sep={\,}}